\documentclass[letterpaper, 10 pt, conference]{ieeeconf}  % Comment this line out if you need a4paper

\IEEEoverridecommandlockouts                              % This command is only needed if 
\usepackage{lineno}
\usepackage{amsmath,amsfonts,amssymb}
\DeclareMathOperator*{\argmaxA}{argmax}
\usepackage{bm}
\usepackage{graphics}
\usepackage{graphicx, subfigure}
\usepackage{caption}

\usepackage{ifoddpage}

\usepackage{adjustbox}
\usepackage{dblfloatfix}
\usepackage{lipsum}
\usepackage{epsfig}
\graphicspath{{Figures/}}
\usepackage{epstopdf}
\usepackage{upgreek}
\newsavebox{\bigimage}

\usepackage{dblfloatfix}

\usepackage{verbatim}
\usepackage{color,soul}

\usepackage[linesnumbered,ruled,vlined]{algorithm2e}
\SetKwRepeat{Do}{do}{while}%
\SetKwInOut{Input}{Input}
\SetKwInOut{Output}{Output\,}
\SetKwInOut{Data}{Data}
\SetKwProg{Tree}{do}{}{while}

\usepackage{colortbl}
\usepackage[noadjust]{cite}
\usepackage{placeins}
\usepackage{lipsum}

\title{\LARGE \bf
Coverage Sampling Planner for UAV-enabled Environmental Exploration and Field Mapping
}

\author{Teng Li$^{1}$, Chaoqun Wang$^{2}$, Max Q.-H. Meng$^{2}$, and Clarence W. de Silva$^{1}$% <-this % stops a space
\thanks{$^{1}$Teng Li and Clarence W. de Silva are with Department of Mechanical Engineering, University of British Columbia, BC V6T 1Z4, Canada
        {\tt\small \{tengli, desilva\}@mech.ubc.ca}}%
\thanks{$^{2}$Chaoqun Wang and Max Q. -H. Meng are with the Department of Electronic Engineering, The Chinese University of Hong Kong, Shatin, Hong Kong
        {\tt\small \{cqwang, qhmeng\}@ee.cuhk.edu.hk}}%
}

\begin{document}

\maketitle
\thispagestyle{empty}
\pagestyle{empty}

%%%%%%%%%%%%%%%%%%%%%%%%%%%%%%%%%%%%%%%%%%%%%%%%%%%%%%%%%%%%%%%%%%%%%%%%%%%%%%%%
\begin{abstract} 
Unmanned Aerial Vehicles (UAVs) have been implemented for environmental monitoring by using their capabilities of mobile sensing, autonomous navigation, and remote operation. However, in real-world applications, the limitations of on-board resources (e.g., power supply) of UAVs will constrain the coverage of the monitored area and the number of the acquired samples, which will hinder the performance of field estimation and mapping. Therefore, the issue of constrained resources calls for an efficient sampling planner to schedule UAV-based sensing tasks in environmental monitoring. This paper presents a mission planner of coverage sampling and path planning for a UAV-enabled mobile sensor to effectively explore and map an unknown environment that is modeled as a random field. The proposed planner can generate a coverage path with an optimal coverage density for exploratory sampling, and the associated energy cost is subjected to a power supply constraint. The performance of the developed framework is evaluated and compared with the existing state-of-the-art algorithms, using a real-world dataset that is collected from an environmental monitoring program as well as physical field experiments. The experimental results illustrate the reliability and accuracy of the presented coverage sampling planner in a prior survey for environmental exploration and field mapping.
\end{abstract}

%%%%%%%%%%%%%%%%%%%%%%%%%%%%%%%%%%%%%%%%%%%%%%%%%%%%%%%%%%%%%%%%%%%%%%%%%%%%%%%%
\section{INTRODUCTION}
Automated sensing systems that can carry out in-situ sampling, real-time processing, and remote deployment have been widely deployed in environmental monitoring programs. Although sensing systems with static sensors have been widely implemented to provide on-line measurements (e.g., \cite{fang2014integrated, mois2017analysis}), their applications have been impeded by the inadequacy and inflexibility in area surveillance. In contrast, unmanned aerial vehicles (UAVs) can offer flexibility, efficiency, and effectiveness in information gathering on the spatial scale. %, which can provide more useful observations for field estimation and spatial analysis.
Consequently, UAV-enabled robotic sensors are being implemented in exploring and monitoring of environmental processes \cite{dunbabin2012robots}.

Depending on the monitoring goals, a UAV is equipped with appropriate sensors to measure the necessary variables/parameters (physical, chemical, biological, etc.). A specific type of sensor may be able to acquire data from multiple geographical locations simultaneously. For example, a thermal camera can capture the temperature distribution over a large spatial area. Many other types of sensors, however, may only acquire data from a single location at one time. For example, an air quality probe measures the target parameter concentration (e.g., PM$_{2.5}$, CO, or SO$_2$  \cite{yang2018real}) at just one site at a time. These sensors are termed ``point sensors'' in the present paper. The measurements from them can provide the information pertaining to the corresponding sampling locations only. Then, to obtain the spatial information, the mobile sensing robots with point sensors have to navigate to sampling locations in the study area to take necessary measurements. The observations made in this manner are utilized to estimate the spatial field or predict the physical quantity of any unobserved locations.

Exploring and mapping an area of interest with point sensors have been actively studied in applications such as geostatistics, robotics, wireless sensor networks, and environmental monitoring. Observations from the sampled sites are utilized to characterize the spatial profile, estimate the underlying environmental model, predict the information at unobserved locations, or reconstruct the scalar field of the monitored area. Sites to be measured are generated following a sampling design, which affects the estimation and mapping performance. To investigate an unknown field of interest, a preliminary coverage sampling phase is required through a prior survey, which is called an exploratory survey phase \cite{muller2007collecting}. A coverage sampling design is required that can distribute plots densely and evenly over the measured area. Subsequently, knowledge-based strategies and implementations can be further performed by using the gathered information.

In a UAV-enabled mobile sensing process, a robot navigates to different locations for carrying out automated sampling and data acquisition. It is evident that sampling at a higher resolution that covers the overall spatial field will provide estimations that are closer to the ground truth of the underlying environmental phenomenon. However, UAVs in the field have limited on-board resources, especially limited energy storage and power supply (batter, fuel, etc.), which restrict the number of collected data samples as well as the associated area of coverage. Accordingly, the major challenge for UAV-based sensing applications resides on how to schedule an effective and efficient sampling mission to visit and measure at different sites over the monitored area. 

This paper presents a sampling planner for UAV-based mobile sensing that integrates both coverage sampling design and associated path planning for exploration and mapping of an unknown environment. The proposed scheduling framework generates a hexagonal grid-based sampling design that provides spatially balanced sampling sites and the sampling path cycle to visit them. The planned sampling mission results in an optimal coverage density for exploratory sampling, subject to a limited power supply budget. The robotic sensor travels along the assigned sampling path to collect data samples. Subsequently, the data is used to estimate a statistical model of the underlying environmental phenomenon in the form of a random field represented by a Gaussian process (GP) with a spatial trend. A robust Kriging method is utilized to estimate the GP structure and construct the scalar map of the monitored parameter. The experimental results are presented and compared with the existing approaches, while highlighting the superior performances of the proposed coverage sampling planner. % and the computational cost. 
% The developments in the present work can be implemented in UAV-enabled mobile sensing systems with point sensors for environmental field exploration and mapping. 
The rest of the paper is organized as follows. Section II introduces the related work. In Section III, the problem formulation of the present work is addressed. The proposed mission planner for coverage sampling and path planning is detailed in Section IV. Section V presents the experiments and discusses the experimental results. The final section concludes the paper.

\section{Related Work}
\label{sec: Related Work}
Environmental field estimation and mapping through mobile sensing is an active research topic. 
% Although model-free approaches have been proven to be effective in applications such as stochastic source seeking \cite{stankovic2010extremum}, a spatial statistical model has attracted attention of researchers for spatial data analysis. 
Model-based approaches have attracted attention of researchers for spatial data analysis since they can estimate the underlying structure of the data generating process and provide the spatial characteristics and global information of the environment. 
% The study area is typically modeled as a random field through a stochastic process. 
% GP model, a nonparametric Bayesian scheme, has been widely studied as the spatial statistical structure to model the underlying field and provide estimation, prediction, and mapping of the monitored area \cite{liu2018cope}. 
For example, by generalizing a Gaussian distribution in a finite vector space, to a Gaussian process in a function space of infinite dimension, the GP framework can be used to model various physical phenomena and estimate values at any location in the sensing domain \cite{xu2015bayesian}.

A distribution of sampling locations provides a pattern across the study area, which indicates the estimation performance of the underlying field. Many frameworks have been proposed to select the target locations for constructing the scalar field. Among them, the experimental design theory has been studied to find the optimum sensor deployment plan by optimizing an information gain that is determined by using an established model. In the research, strategies were addressed to select the locations with high ``informativeness'' to achieve adaptive sensing (e.g., \cite{nguyen2017adaptive}). However, these strategies rely on prior knowledge of the environmental model structure. Deviations of the assumptions will lead to model misspecification and unacceptable estimation performance. In other words, the field estimation results will not be robust against misspecification of the established model. In many circumstances, the assumptions may not be practically feasible. For instance, the assumption of a known and constant mean of the environmental model (e.g., \cite{evans2016environmental}) can cause inferior estimation performance in a practical situation where unknown spatial trends exist.

For an unknown environment, the underlying environmental model is learned by utilizing observations that are taken across the study area. 
%The approximation of the environmental field includes the components of the model structure, such as model order, basis function, and hyperparameters. 
Coverage sampling that is done to relax unrealistic assumptions is called an exploratory design, which provides sufficient flexibility to explore an unknown field \cite{muller2007collecting}. Such designs have been adopted in spatial mapping without rigid prior assumptions, for initial collection of prior knowledge, exploratory survey for long-term design improvement, or combination with optimum designs for improved estimation performance. In most existing monitoring programs, the exploratory sampling is required to learn the underlying field. Regular grid-based sampling approaches accomplish such objectives as exploratory designs \cite{paull2013sensor, de2015geospatial}. These approaches distribute sampling sites by decomposing the field into a grid of cells (e.g., squares, triangles, hexagons) where the samples are taken in the grid tessellation. In the absence of prior knowledge, a regular grid ensures overall coverage of the surveillance area. In particular, it provides a simple frame to distribute spatially balanced data samples across the study area for prior survey.

For carrying out a sampling mission with a robotic sensor, a path to visit target sites is planned. The traveling salesman problem (TSP) \cite{applegate2006traveling} has been widely applied by connecting all the target locations directly. It finds the path by visiting each location exactly once and leads to the shortest route for mobile sensing. However, the TSP solvers are nondeterministic polynomial-time hard (NP-hard) \cite{gutin2006traveling}; thus the computational cost is extremely high for a large scale problem. Most critically, these methods do not provide freedom to iteratively adjust the sampling pattern for better estimation performance while ensuring that the cost of the resultant path does not exceed the available power. % Rather than connecting the target points to form a sampling path, 
In contract, many other planners try to generate a coverage path first and then plot the samples along the path. In the literature, cellular decomposition methods, such as Boustrophedon path \cite{fang2010coverage}, lawnmower path \cite{somers2016human}, and their variants \cite{mora2013analysis, wilson2017adaptive}, have been studied to generate a transect survey pattern consisting of a series of parallel linear transects to cover the study regions. Then data samples are taken while traveling back and forth along the generated coverage path. The work in \cite{galceran2013survey} reviews the strategies that can plan a coverage path. Although these planners can generate coverage paths, most of them focus on the problem of sweeping over the area without analyzing the effect of the resulting sampling frames on spatial field mapping. Additionally, a constrained total travel length has not been addressed or emphasized in these planners. 

To the best of our knowledge, scheduling of a UAV-enabled robotic sensor under an energy constraint in field exploration and mapping has not been adequately investigated. The present work introduces a scheme that integrates both coverage sampling design and path planning, which guides the sensing agent to take measurements for estimating and mapping an unknown environment more effectively. The proposed method can plan a path cycle with an optimal coverage density for sampling under a power supply budget.

\section{Problem Formulation}
\label{sec: problem formulation}
\subsection{Random Field \& Environmental Model}
Environmental phenomena are reflected by complex interactions of ecological processes. The variation of an environmental property appears to be random, which is commonly modeled as a random variable in spatial statistics. A set of random variables can be represented as a random field. The associated nomenclature of the present paper is given as follows. The region of interest is a 2-dimensional plane that is denoted by the set $A\subset \mathbb{R}^2$ with its contour $\hat{A}$. The continuous plane is discretized as a set of locations $\tilde{S}$ with the possible sampling locations. Let $\mathbf{s}\in \mathbb{R}^2=(s_1, s_2)$ denote a 2-dimensional coordinate vector of a sampling location in the surveyed field. A random field $\mathbf{Y(s)}=\{Y(\mathbf{s}), \mathbf{s}\in \tilde{S} \}$ is a set of random variables $Y(\mathbf{s})$ indexed by its associated locations $\mathbf{s} \in \tilde{S}$. A physical quantity can be observed from a random variable $Y(\mathbf{s})$ at its sampled location. In the present work, the random variable is represented by the model:
\begin{equation}
\label{eq1: environmentDataModel}
Y(\mathbf{s}) = X(\mathbf{s})+Z(\mathbf{s}),
\end{equation}
where $X(\mathbf{s})$ is a regression function that defines the mean value of the process (spatial trend over space) and $Z(\mathbf{s})$ is a stochastic function that defines the random variation. Consider the finite random variables at the locations in $\tilde{S}$, the model in Equation (1) is vectorized as:
\begin{equation}
\label{eq1: environmentDataModelFinal}
\mathbf{Y}(\mathbf{s}) = \mathbf{X}(\mathbf{s})+\mathbf{Z}(\mathbf{s}).
\end{equation}
% a Gaussian process of zero mean, variance $\sigma^2$ and covariance function $\mathbf{C}(\mathbf{s},\mathbf{s}')$.
Equation (2) represents the underlying environmental model.

To study the random field $\mathbf{Y(s)}$, it is required to take finite realizations at some sampled locations and make inferences on other unobserved random variables using these data samples and the environmental model. Given any set of $\mathcal{N}$ sampled locations $S= \{\mathbf{s}_1, ..., \mathbf{s}_\mathcal{N}\}$,  the corresponding realizations are denoted as $\mathbf{y(s)} = [y(\mathbf{s}_1),...,y(\mathbf{s}_\mathcal{N})]^\intercal$, referring to the set of observations at these locations.

% where $\mathbf{Y}(\mathbf{s}) = [Y(\mathbf{s}_1), ..., Y(\mathbf{s}_\mathcal{N})]^\intercal, \mathbf{X}(\mathbf{s}) = [X(\mathbf{s}_1),...,$ $X(\mathbf{s}_\mathcal{N})]^\intercal$, and $\mathbf{Z}(\mathbf{s}) = [Z(\mathbf{s}_1), ..., Z(\mathbf{s}_\mathcal{N})]^\intercal$ are random vectors. 

\subsection{Sampling Design \& Mobile Sensing Model}
Statistical model and its characteristics of a random field are interpreted according to the distribution of sampling locations over the field. In the present work, a hexagonal grid framework is used to generate the locations of interest for coverage sampling and plan the associated coverage path. Specifically, the sensing domain of the continuous planar area $A$ is first decomposed into a grid of hexagonal cells. The cells that have their centroids within the study area $\hat{A}$ are designated as the sub-regions of interest (SRoIs). These cell centroids are the sampling locations of interest (SLoIs), $\mathbf{s}\in S$, for collecting data samples. An execution example of the proposed design of coverage sampling in a hexagonal tessellation is shown in Fig. 1(a).

In the sampling design, $l$ denotes the edge length of a hexagon in the grid; and $SL$ and $|SL|$ denote the point set and the total number of the generated SLoIs, respectively. In the context of mobile sensing with a UAV, let the sequence $p = ({\bf{s}}_1, {\bf{s}}_2, ..., {\bf{s}}_\mathcal{N})$ represent a planned path, where ${\bf{s}}_n\in SL, n=1,2,...,\mathcal{N}$ denotes a sampling location (waypoint) that is visited within the path $p$. $|p|$ denotes the total path length. The UAV visits and operates the measuring process at the locations by following the order in the sequence. Let the set $SP$ and $|SP|$ represent the set and the number of sampling locations that are visited along this path, respectively. Let $d_{{\bf{s}}_i, {\bf{s}}_j}$ and $e(d_{{\bf{s}}_i, {\bf{s}}_j})$ represent the distance and energy cost when traveling from location ${\bf{s}}_i$ to ${\bf{s}}_j$, respectively. In the measuring process, the UAV hovers at a SLoI to carry out the in-situ measurement using a point sensor. Let $e_M$ represent the energy consumption of the entire measurement process at a SLoI. The total energy cost of traveling along the path $p$ is expressed as:
\begin{equation}
e(p)=\sum_{i=1}^{\mathcal{N}-1} e(d_{{\bf{s}}_i, {\bf{s}}_{i+1}})+e_{M}\cdot |SP|, {\bf{s}}_n\in SP\subseteq SL.
\end{equation}

The objective of the present planner is to generate a sampling mission $p = ({\bf{s}}_1, {\bf{s}}_2, ..., {\bf{s}}_\mathcal{N})$ for scheduling a UAV-enabled sensor to collect data samples $\mathbf{y}$ at the target locations in $SL$ with the densest sampling resolution under the power supply budget $e_{bdt}$ of the UAV. The data acquired in this manner can be utilized to estimate the underlying field model $\mathbf{Y(s)}$ and map the scalar field $A$. The objective is formulated as:
\begin{equation}
\begin{split}
p^*=\argmaxA_p |SP|, SP\subseteq SL, \text{s.t. } e(p) \leq e_{bdt}.
\end{split}
\end{equation}

\section{Coverage Sampling Planner}
\label{sec: sampling mission planner}
% In the designed coverage sampling frame, the sensing domain is decomposed into SRoIs with spatially-balanced SLoIs that are distributed over the study area. 
% In this section, a hexagonal grid-based coverage path planner that visits the designed sampling locations is presented.
% It generates a path cycle that can guide the robotic sensor to measure the largest number of SLoIs under the energy budget.

%Specifically, the path is constructed to visit the largest number of SLoIs, resulting in a path cycle that visits each location only once and returns to its starting position without exceeding the available energy budget.

\subsection{Coverage Path Planning}
To carry out the sampling design, an ordered sequence is generated that indicates the sequential SLoIs to be visited. A coverage path is planned to visit each SLoI in the order of the sequence. This task is carried out by finding the effective edges (path segments) between the neighboring SLoI pairs to form the final path cycle. The edge candidates between the adjacent SLoIs compose a triangular grid pattern (see Fig. 1(b)). The planning goal is to determine the effective edges among the candidates to form a path cycle that visits each location only once and ends at the starting position. % This problem is similar to the formulation of a Hamiltonian cycle on a specified triangular grid graph \cite{gordon2008hamiltonian}. 

To determine the effective edges, an auxiliary coarse cell (ACC) decomposition method is proposed. In this decomposition, each coarse cell contains four neighboring regular hexagons (fine cells). The target edges are constructed by examining the number and the positions of the SLoIs within and surrounding an ACC. An execution example of the edge candidates and the ACC decomposition is shown in Fig. 1(b).

    	\begin{figure}[b!]
    \centering
       \subfigure[]{
    \includegraphics[width=0.22\textwidth]{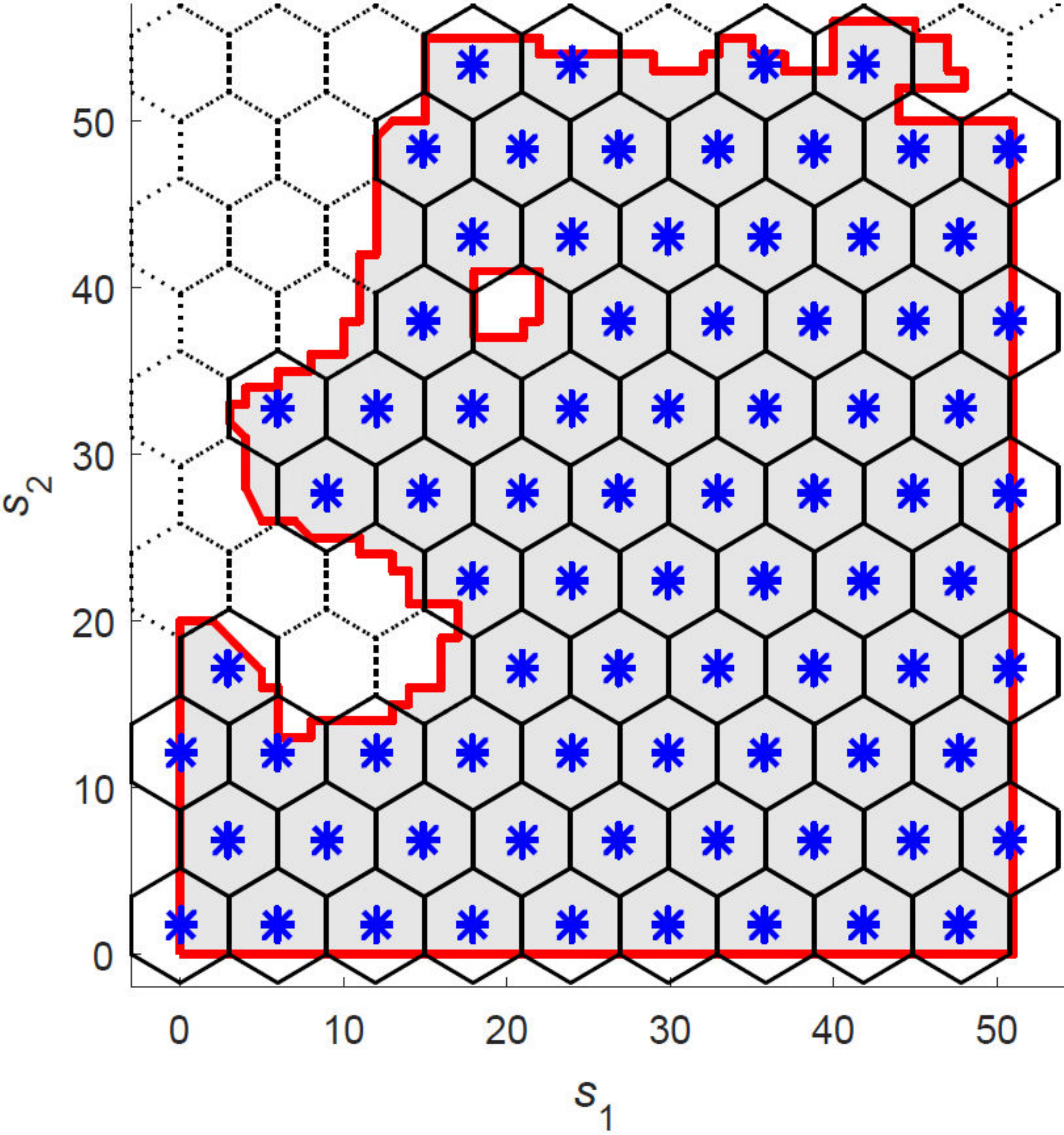}} \label{fig: sampling_design}
        \subfigure[]{
    \includegraphics[width=0.22\textwidth]{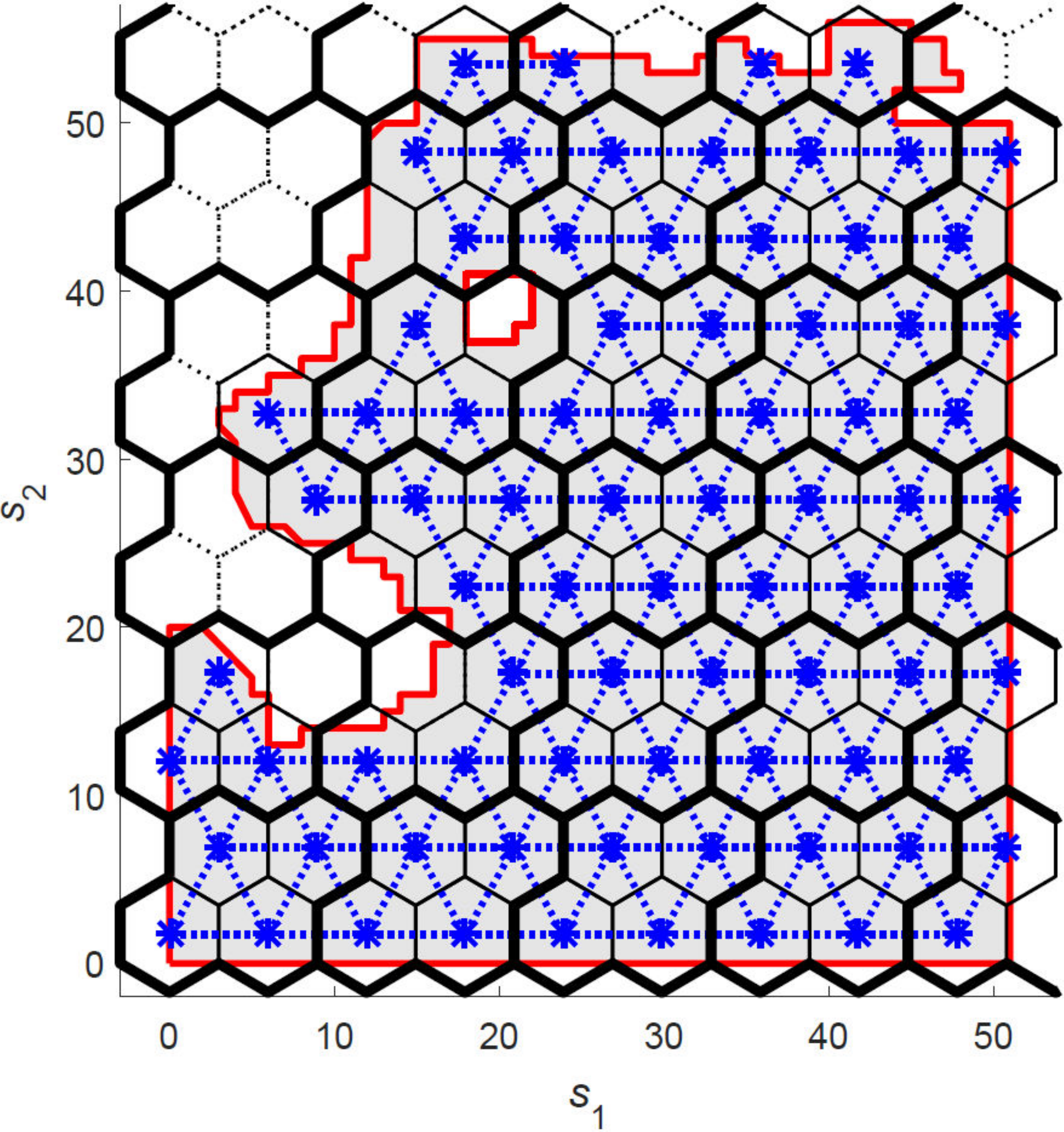}} \label{fig: edge_candidates}
    \caption{(a) Coverage sampling design. The solid red line denotes the study area contour $\hat{A}$. The blue stars denote the SLoIs. (b) Edge candidates and ACC decomposition. The dashed blue lines between the SLoIs denote the edge candidates. The bold black lines denote the ACC contours.}
\end{figure}

\newpage   
The proposed planner traverses the ACCs one by one to generate the path segments depending on the conditions (number and position) of the SLoIs in the current ACC and its surrounding ACCs. In a coarse cell, the bottom left, bottom right, top left, and top right hexagonal polygons are designated as the fine-cells 1, 2, 3, and 4, respectively. The current coarse cell is considered as a full cell if it contains four fine cells. The neighboring four fine cells in its right and bottom neighboring coarse cells are defined as the cells 5, 6 and cells 7, 8, respectively. The related SLoIs of an ACC for the path segment generation is shown in Fig. 2(a). In the figure, the SLoIs of the current and its neighboring coarse cells are labeled with the corresponding numbers. The dashed blue lines show the possible path segments to be generated. When traversing ACCs, for a selected coarse cell that is full, the segment paths are constructed by the following rules:
\begin{itemize}
\item 	If the left neighboring ACC is full, generate the edge $e_{3,4}$; otherwise, generate the edges $e_{1,3}$ and $e_{3,4}$.
\item If the right neighboring ACC is full, generate the edges $e_{2,5}$ and $e_{4,6}$; otherwise, generate the edge $e_{2,4}$.
\item If the bottom ACC is full and the current ACC row has not been connected to the bottom ACC row, generate the edges $e_{1,7}$ and $e_{2,8}$, and remove the existing edge $e_{7,8}$; otherwise, generate the edge $e_{1,2}$.
\end{itemize}

After executing these procedures, a path cycle is generated to visit all the full ACCs. However, there are still unvisited SLoIs within the area contour whose ACCs are not full. To visit them, several strategies for path segment generation are applied. First of all, for an unvisited SLoI, if there is a pair of neighboring SLoIs with a constructed edge between them, add two new edges connecting the current SLoI to the two neighbors, and then remove the edge between the neighbors (see Fig. 2(b)). In addition, if there are four neighboring SLoIs with two possible ``Z" shaped patterns in Fig. 2(c), then the modification rules of the path segments are shown in the figure. The above-mentioned strategies are referred as the $V$- and $Z$-modifications in the work of Arkin et al. \cite{arkin2009not}. The coverage path is updated by iteratively checking and implementing these strategies on all unvisited SLoIs. It is noted that the planned results may form more than one path cycle. Then, any two neighboring cycles can be combined to form one cycle by a simple edge adjustment (see Fig. 2(d)).

When compared with the work in \cite{arkin2009not}, the proposed scheme is more robust for dealing with a triangular grid graph. The compared work requires that the planner starts from a boundary cycle of a 2-connected polygonal triangular grid. In contrast, the proposed method tackles the path cycle from inside to outside. There are no specific requirements on the characteristics of the input triangular grid graph. The proposed hexagonal grid-based coverage (HGC) sampling planner is summarized by the pseudo code in Algorithm 1. The code from line 1 to 21 generates the coverage path to visit all full ACCs. The code from line 22 to 32 deals with the unvisited ACCs. The algorithm generates a coverage path cycle to travel among the expected sampling sites. Fig. 3 shows the execution examples of the proposed method.
% The data samples are acquired by following the generated coverage path with equally spaced distribution.
       \begin{figure}[!h]
      \centering
      \subfigure[Related SLoIs and paths.]{\label{fig:3a}  
      \includegraphics[width=0.20\textwidth]{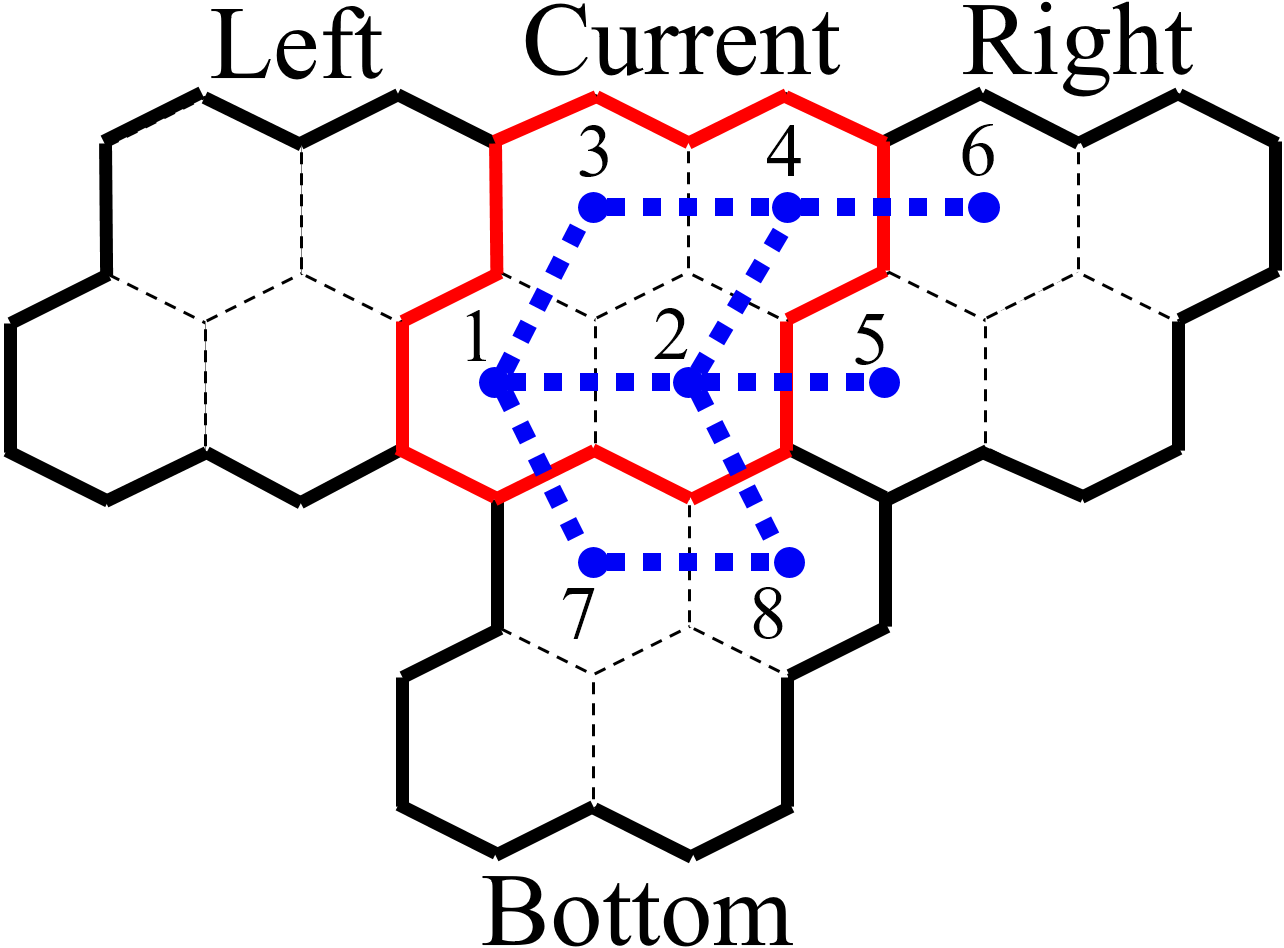}} \
      \subfigure[\textit{V}-modification.]{\label{fig:3b}  
      \includegraphics[width=0.23\textwidth]{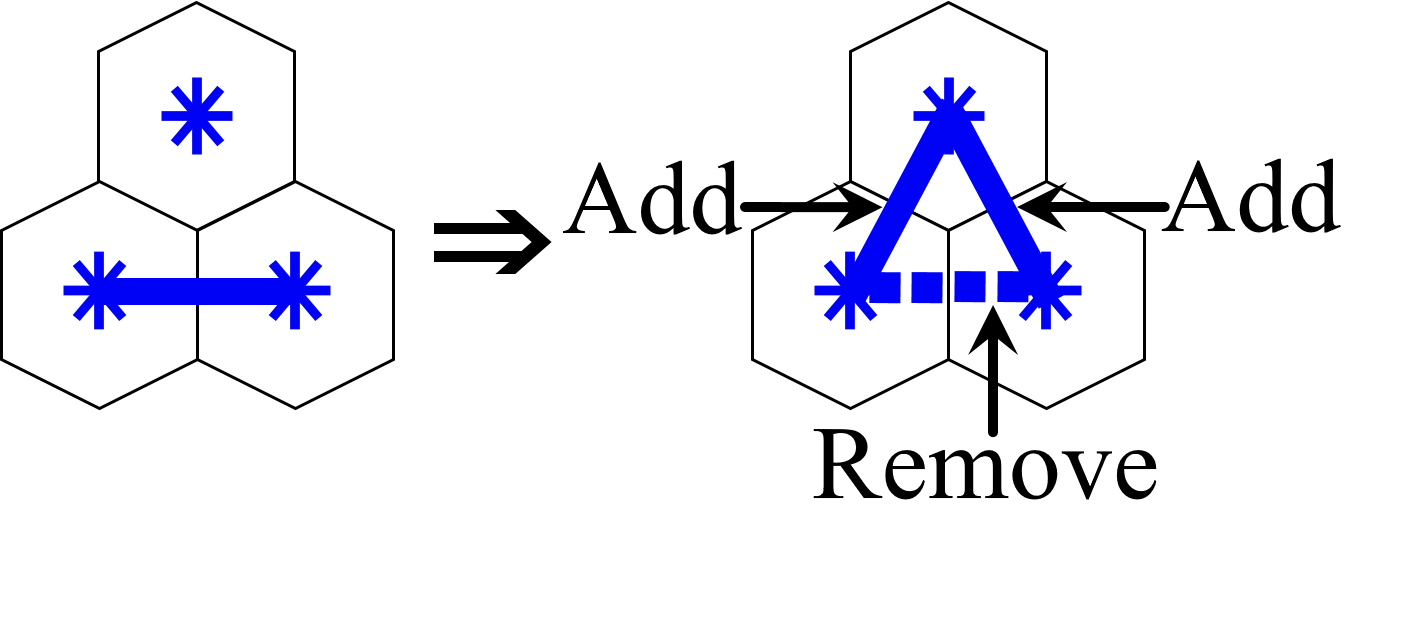}} \\
      \subfigure[\textit{Z}-modification.]{\label{fig:3c} 
      \includegraphics[width=0.20\textwidth]{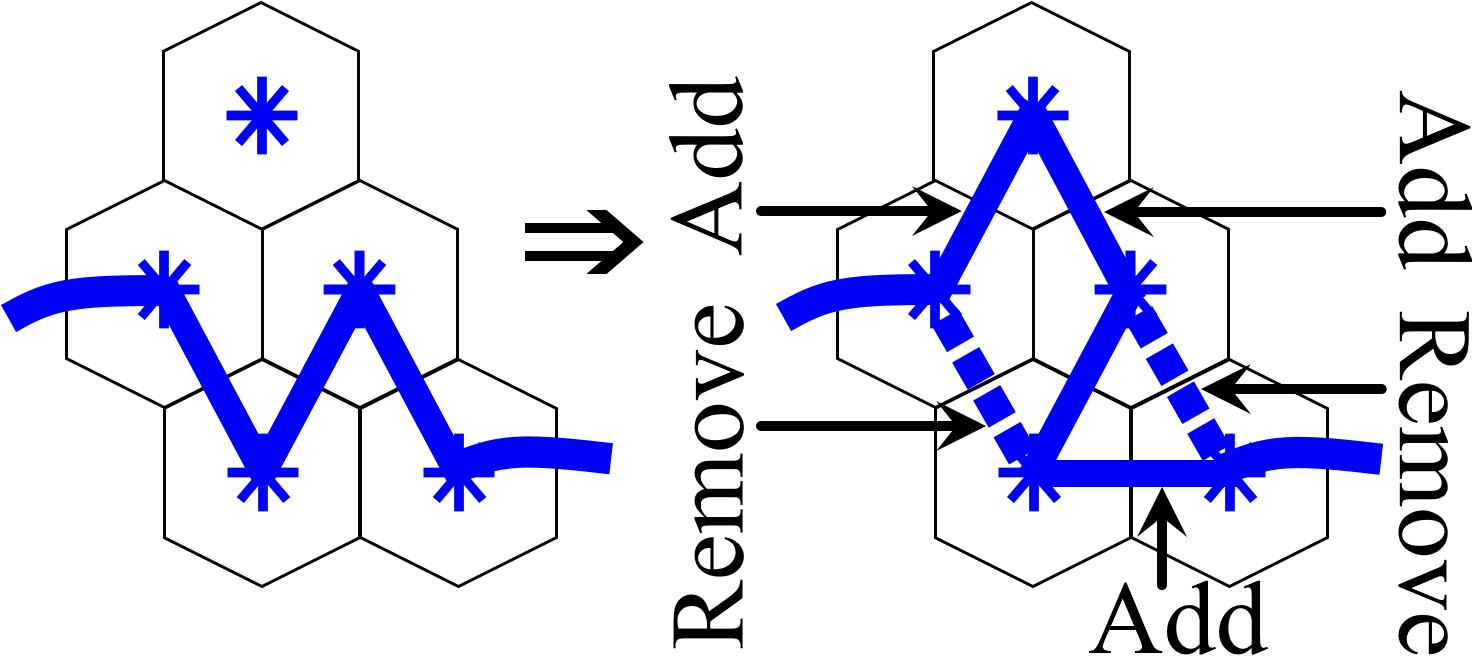} \ \ \
      \includegraphics[width=0.22\textwidth]{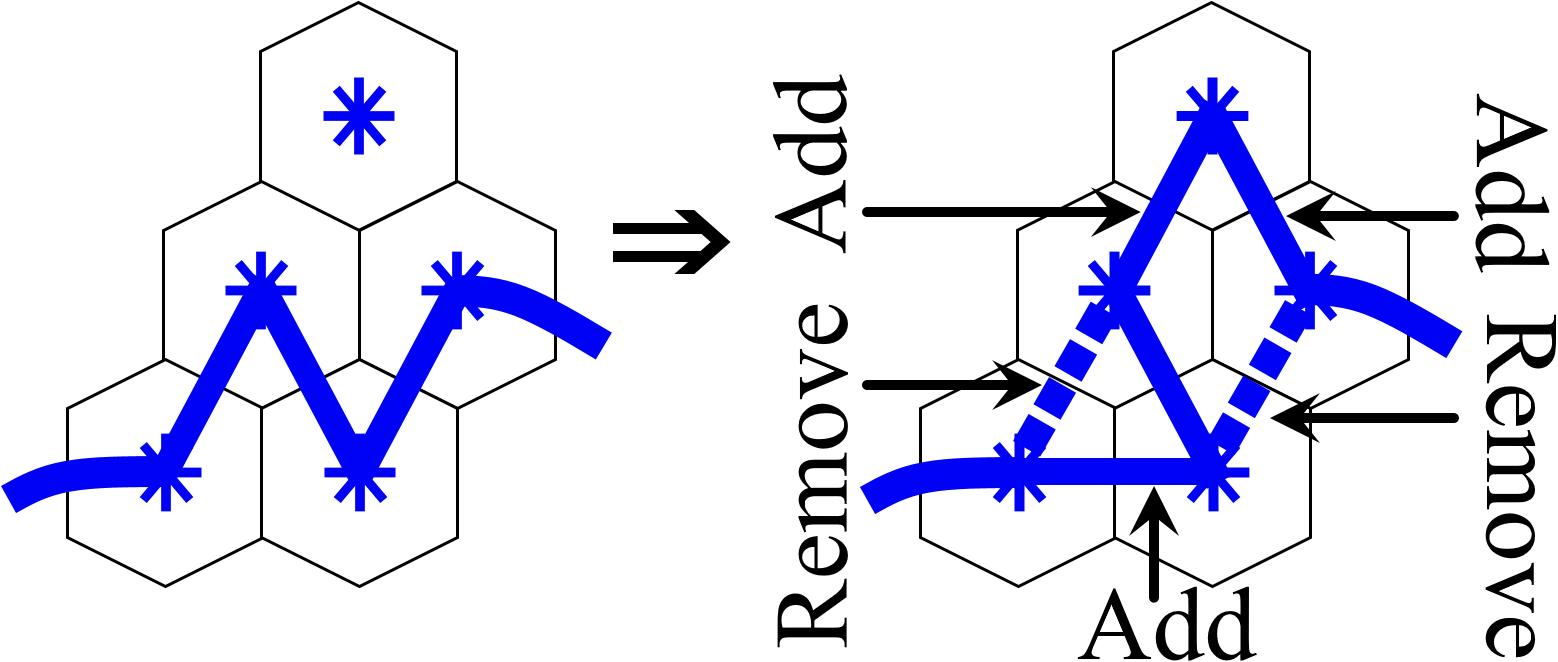}} \\
      \subfigure[Cycle combination.]{\label{fig:3d}  
      \includegraphics[width=0.35\textwidth]{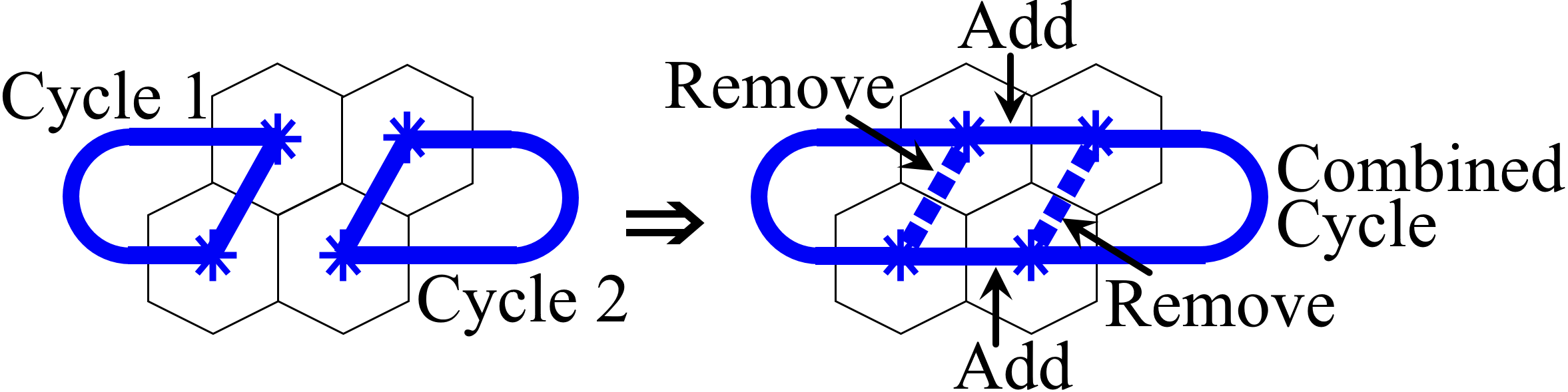}}

          \caption{Strategies of path segment generation.}
          % (a) Path segment generation in an ACC. The current ACC is highlighted by the solid bold contour. The SLoIs of the current and its neighboring ACCs are labeled by their numbers. The dashed blue lines show the possible path segments to be generated. (b) Path cycle combination of two neighboring cycles; (c) Path segment generation for unvisited SLoIs.
      \label{figurelabel}
   \end{figure}

  	   \begin{figure}[!h]
      \centering
      \subfigure[Visiting full ACCs.]{\label{fig:4a}  
      \includegraphics[height=4.2cm]{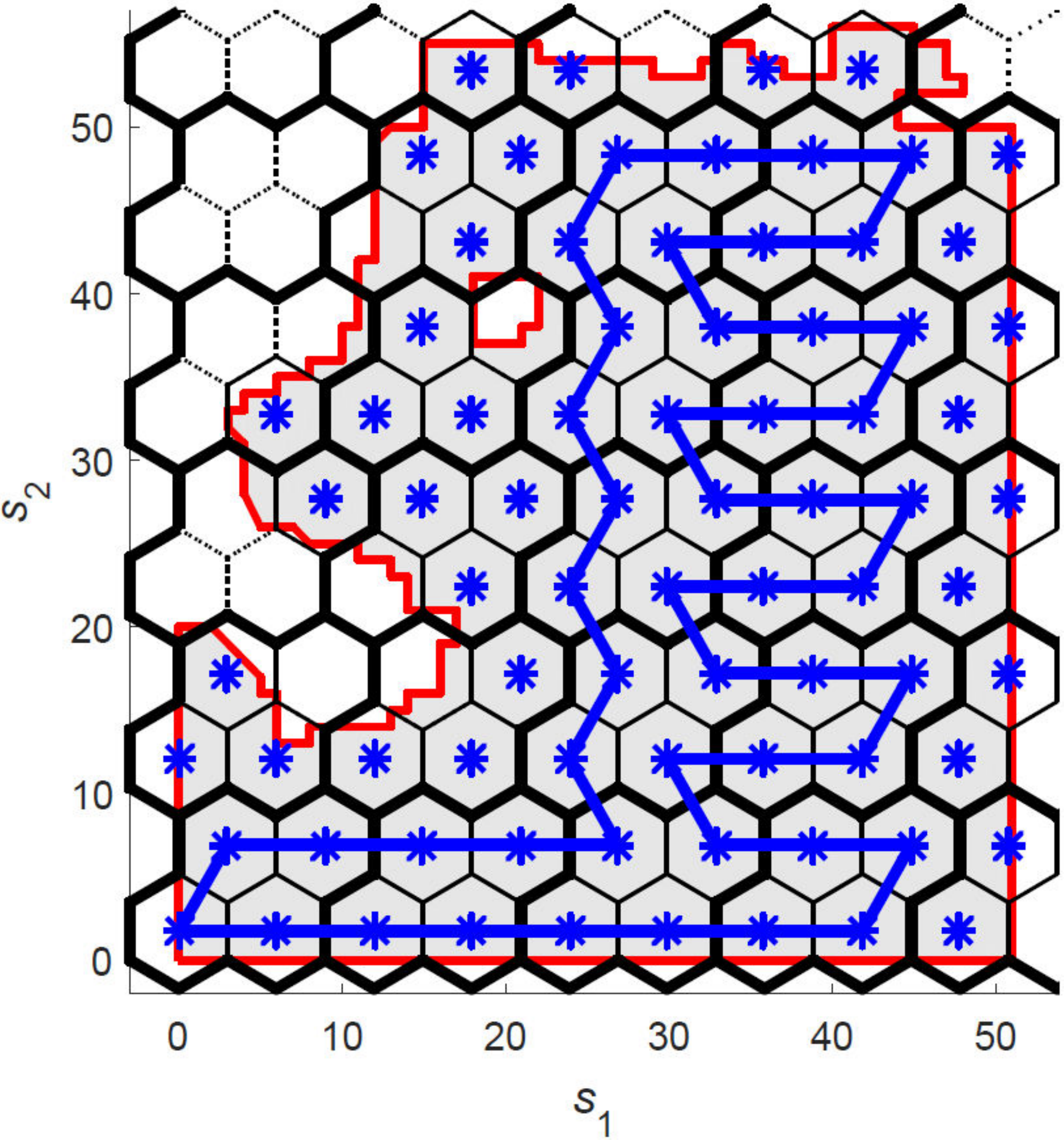}} %\ \ \
      \subfigure[Incorporating unvisited SLoIs.]{\label{fig:4b}  
      \includegraphics[height=4.2cm]{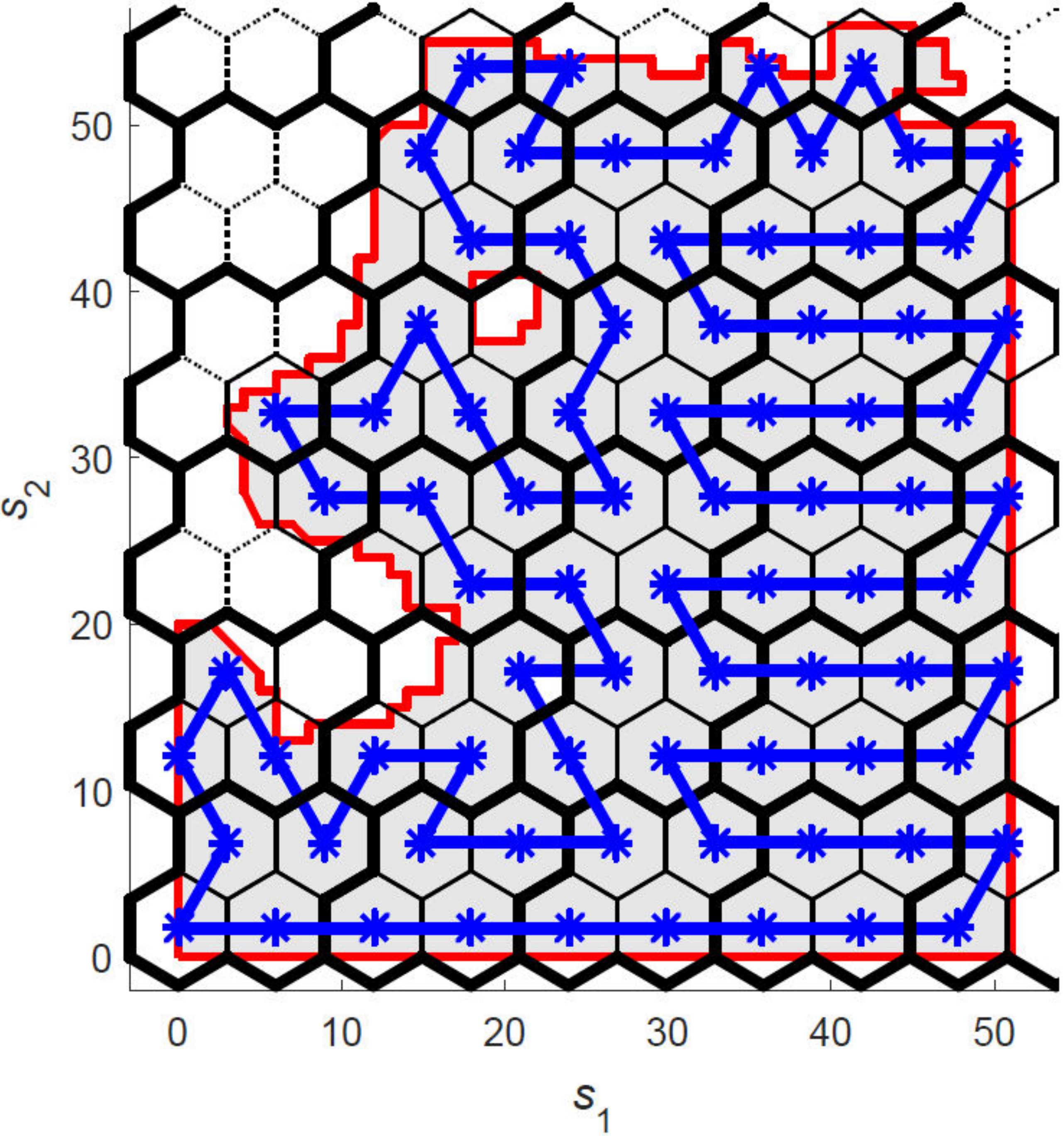}} %\ \ \
	  % \subfigure[]{\label{fig:4c}	   
	  % \includegraphics[height=4.5cm]{Fig4c.pdf}} % \ \ \
\caption{Execution examples of the HGC sampling planner.} %; (c) final coverage path for sampling.}
          % (a) The path segment generation to visit full ACCs. The solid blue lines between APs denote the selected edges as the generated path segments. (b) The generated coverage path after incorporating all unvisited SLoIs. The solid blue lines form the cycle of the planed coverage path. (c) The final coverage path for sampling. The blue star markers denote the distributed sampling locations along the path.
      \label{fig:4}
   \end{figure}

\subsection{Optimal Sampling Density}
The distances between any neighboring SLoI pairs are equal, which indicates the spatial sampling resolution. Given an edge length $l$ of the fine cells in the hexagonal grid, the distance is given by $d = \sqrt{3}\cdot l$. For the area contour $\hat{A}$, the total number of SLoIs to by surveyed, $|SL|$, can be determined as a dependent variable of $l$ as well as $d$. In view of the equal distances between SLoI pairs, the total path length $|p|$ of the generated coverage path $p$ can be determined as $|p|=d\cdot |SP|$, $|SP|\leq |SL|$. Given an energy budget $e_{bdt}$, finding the maximum point set $SP$, $SP\subseteq SL$, that the coverage path can achieve provides the most dense coverage sampling and further improves the exploration performance of field estimation and mapping. The optimal sampling density $d^*$ is determined by the optimal edge length $l^*$ as:
\begin{equation}
\label{eq4: optimalR}
\begin{split}
& l^* = \argmaxA_l |SP|,  SP\subseteq SL, \\ 
& \text{s.t.}  \ |SP| \leq |SL|, \\
& \ \ \ \ \ e(p) = e(\sqrt{3}\cdot l) \cdot |SP| + e_M\cdot |SP| \leq e_{bdt}.
\end{split}
\end{equation}

A decreasing grid spacing $d$ intuitively increases the resultant number of SLoIs, $|SL|$, inside the study area. However, there is no analytical expression for determining $|SL|$ given $l$, since the shape of the study area is generally irregular and complex. The dependent variable $|SL|$ has to be determined by executing the ACC decomposition for a given $l$. The optimal $d^*=\sqrt{3}\cdot l^*$ is chosen as the optimal coverage density of sampling to generate the final path. The efficiency of finding the optimal $d^*$ can be operated by using the binary search to determine $l^*$ such that  $0\leq e_{bdt}-e(p)\leq \delta$.
% A brute-force search can be utilized to find the $d^*$ that satisfies Equation (\ref{eq4: optimalR}), e.g., $d^* = \argmaxA_d |SP|, d\in [d_{\min}:\Delta d:d_{\max}]$.

The proposed planner guarantees that the total length of the generated path is bounded and indicative referring to the cell size. There is no need to execute the path generation algorithm before determining the optimal edge length $l$. It considerably reduces the computational complexity and the corresponding processing time when finding the optimal coverage density. The coverage sampling path can be obtained by the planner in Algorithm 1. The proposed hexagonal grid-based coverage (HGC) sampling planner is summarized by the pseudo code in Algorithm 2. 

\SetInd{0.35em}{0.35em}
% \SetAlgoNoLine
\DontPrintSemicolon
\begin{algorithm}[!t]
\small
  \caption{genCoveragePathCycle}\label{alg1}
  \Input{$\hat{A}, d, {\bf{s}}_1\in S$.}
  \Output{$p=({\bf{s}}_1,{\bf{s}}_2,\ldots,{\bf{s}}_\mathcal{N})$.}
  $colACC, rowACC = getACCHexagonalGrid(\hat{A}, d)$\\
  \For {$i = 1:colACC$}
  {
    \For {$j = 1:rowACC$}
    {
      $SLoIs=getCurCellLocation(i,j)$\\
      \eIf {$isCurrentACCFull(SLoIs)$}
      {
        \If {$\sim isNearACCFull($\upshape{`LEFT'}$)$}
          {$p \leftarrow genPath(e_{1,3})$\\
           $flag = \text{True}$\\}
        \eIf {$isNearACCFull($\upshape{`RIGHT'}$)$}
          {$p \leftarrow genPath(e_{2,5},e_{4,6},e_{3,4})$\\}
          {% else
            $p \leftarrow genPath(e_{2,4},e_{3,4})$\\
          }
        \eIf {$isNearACCFull($\upshape{`BOTTOM'}$) \ \& \ isTrue(flag)$}
        { $p \leftarrow genPath(e_{1,7},e_{2,8})$\\
          $p \leftarrow removePath(e_{7,8})$\\
          $flag = \text{False}$\;}
        {% else
          $p \leftarrow genPath(e_{1,2})$\\
        }
        $V = setFineCellVisited(SLoIs)$\\
      }
      { % else
        $U = setFineCellNotVisited(SLoIs)$\\
      }
    }
  }
  \If {$getPathCycleNumber(p)>1$}
  {$p \leftarrow pathSegmentGeneration$($p$)\\
  }
  \Do
  {$~isEmpty(updateCells)$}
  {
    $updateCells = \emptyset$\\
    \ForEach {$ i\in SN$}
    {
      $cells, pattern = findNearCellsInPatterns$($i$)\\
      \If {$\sim isEmpty(cells)$}
      { $updateCells \leftarrow updateCells\cup i$\\
        $p \leftarrow pathSegmentGeneration(p, cells,pattern$)\\
        $SP \leftarrow SP\cup i$\\
        $SN \leftarrow SN\setminus i$\\
      }
    }
  }
\end{algorithm}
	\SetInd{0.35em}{0.35em}
% \SetAlgoNoLine
\DontPrintSemicolon
\begin{algorithm}[!h]
  \caption{coverageSamplingMissionPlanner}\label{alg2}
  \Input{$\hat{A}, e_{bdt}, {\bf{s}}_1\in S$}
  \Output{$p=({\bf{s}}_1, {\bf{s}}_2, ... , {\bf{s}}_\mathcal{N})$}
  $l^*\leftarrow searchOptimalCoverageDensity(\hat{A}, e_{bdt})$\\
  $SLoIs \leftarrow genCoverageSamplingDesign(\hat{A}, l^*)$\\
  $p \leftarrow genCoveragePathCycle(\hat{A}, SLoIs, {\bf{s}}_1)$\\
\end{algorithm}

\subsection{Model Estimation \& Field Mapping}
%The path obtained can be implemented in a mobile sensing robot to perform the sampling mission. The robot travels along the path by following the sequential SLoIs and senses at each location. 
The data samples taken along the planned path are used to estimate the underlying environmental model and build the field map.
In the present work, the universal Kriging method \cite{cressie2015statistics} is implemented to estimate the unknown field. 
In the universal Kriging model, the regression function $\mathbf{X(s)}$ of the environmental model in Equation (2) is treated as a multivariate polynomial; that is:
\begin{equation}
  \mathbf{X(s)}=\Sigma_{i=1}^\mathcal{I} a_i f_i(\mathbf{s}) = \mathbf{f(s)}^\intercal\mathbf{a},
\end{equation}
where $\mathbf{f(s)} = [f_1(\mathbf{s}),f_2(\mathbf{s}),...,f_\mathcal{I}(\mathbf{s})]^\intercal$ represents the basis function (e.g., the power base for a polynomial), $\mathbf{a}=(a_1, a_2, ...,a_\mathcal{I})^\intercal$ represents the coefficients of the regression function. Meanwhile, the stochastic function in Equation (2) is modeled as a GP: $\mathbf{Z(s)}\sim \mathcal{GP}(\mathbf{0, C})$ with zero mean, variance $\sigma$, and a covariance function $\mathbf{C}(\mathbf{s}_i,\mathbf{s}_j)=\mathbf{C}(\bm{\uptheta})$. The covariance function is also known as a kernel, describes the spatial dependence between the locations ${\mathbf{s}}_i$ and ${\mathbf{s}}_j$, $i\neq j = 1,2,...,\mathcal{N}$, where $\bm{\uptheta}$ denotes the hyperparameters. As a result, the random field $\mathbf{Y(s)}$ can be expressed as a GP: $\mathbf{Y(s)}\sim \mathcal{GP}\big(\mathbf{X(s)}, \mathbf{C}(\mathbf{s}_i,\mathbf{s}_j)\big)$. % of mean defined by $\mathbf{X(s)}$ and stochastic residual defined by $\mathbf{Z(s)}$. 

Given a set of samples $S=\{\mathbf{s}_1,\mathbf{s}_2,...,\mathbf{s}_\mathcal{N}\}$ with the corresponding observations $\mathbf{y} = [y(\mathbf{s}_1),y(\mathbf{s}_2),...,y(\mathbf{s}_\mathcal{N})]^\intercal$, the prediction mean and variance are derived by utilizing the Kriging framework at the locations $\mathbf{s}'\in \tilde{S}\backslash S$ as follows:
\begin{gather}
\hat{Y}(\mathbf{s}')=\bm{\mu}(\mathbf{s}')= \mathbf{X(s')}+\mathbf{c}(\mathbf{s}')^\intercal \mathbf{C}^{-1}(\mathbf{y}-\mathbf{Fa}), \\
\bm{\Sigma}^2(\mathbf{s}') = \sigma^2-\mathbf{c}(\mathbf{s}')^\intercal\mathbf{C}^{-1}\mathbf{c}(\mathbf{s}')
+ \mathbf{M}^\intercal(\mathbf{F}^\intercal\textbf{C}^{-1}\mathbf{F})^{-1}\mathbf{M},
\end{gather}
where $\mathbf{F}=[\mathbf{f}(\mathbf{s}_1)^\intercal, \mathbf{f}(\mathbf{s}_2)^\intercal, ...,\mathbf{f}(\mathbf{s}_\mathcal{N})^\intercal]^\intercal$ is the attribute matrix for $S$,  $\mathbf{a}= (\mathbf{F}^\intercal\mathbf{C}^{-1}\mathbf{F})^{-1}\mathbf{F}^\intercal\mathbf{C}^{-1}\textbf{y}$ is derived by generalized least squares, $\mathbf{c}(\mathbf{s}') = [c(\mathbf{s}',\mathbf{s}_1), c(\mathbf{s}',\mathbf{s}_2), ..., c(\mathbf{s}',\mathbf{s}_\mathcal{N})]^\intercal$ defines the correlations between $\mathbf{s}'$ and the locations in $S$, and $\mathbf{M=f(s}')-\mathbf{F}^\intercal\mathbf{C}^{-1}\mathbf{c(s}')$. % The Kriging model is the Best Linear Unbiased Predictor (BLUP) for estimating and predicting the random scalar field under the Gaussian process scheme \cite{cressie2015statistics}.

Consider rare prior knowledge of the underlying field, the blind Kriging approach \cite{joseph2008blind} is adopted in the present paper to identify the regression function. It makes approximation by extending the general universal Kriging model \cite{cressie2015statistics} with additional candidate functions in the regression analysis. The regression function is designated by unknown basis functions in the multivariate polynomial utilizing a Bayesian feature selection method; that is:
\begin{equation}
\begin{split}
  \mathbf{X(s)} & %= \Sigma_{i=1}^\mathcal{I} a_i f_i(\mathbf{s}) + \Sigma_{j=1}^\mathcal{J} b_j g_j(\mathbf{s}) \\
  = \mathbf{f(s)}^\intercal\mathbf{a} + \mathbf{g(s)}^\intercal\mathbf{b},
\end{split}
\end{equation}
where $\mathbf{g(s)}=[g_1(\mathbf{s}), g_2(\mathbf{s}), ...,g_\mathcal{J}(\mathbf{s})]^\intercal$ denotes the set of $\mathcal{J}$ candidate functions, and $\mathbf{b}=[b_1,b_2,...,b_\mathcal{J}]^\intercal$ denotes the corresponding scores of the candidate functions. %Given random field $\mathbf{Y(s)}$, the corresponding scores $\mathbf{b}$ are considered Gaussian, $\mathbf{b}\sim \mathcal{GP}(\mathbf{0},\tau^2\mathbf{R(C)})$, where $\mathbf{R(C)}$ is defined as the variance-covariance matrix. The posterior estimation of $\mathbf{b}$ can be derived by the BK estimation using the data samples $\mathbf{y}$. 
More details of the blind Kriging method are found in \cite{couckuyt2012blind}. For the stochastic term, the Mat\'ern 5/2 kernal is selected as the covariance function, which is expressed as:
\begin{equation}
  \mathbf{C}(\mathbf{s}_i,\mathbf{s}_j)=\sigma^2(1+\dfrac{\sqrt{5}\cdot h}{\rho}+\dfrac{5\cdot h^2}{3\cdot \rho^2})\exp(-\dfrac{\sqrt{5}\cdot h}{\rho}),
\end{equation}
where $h=||\mathbf{s}_i - \mathbf{s}_j||$ denotes the Euclidean distance between $\mathbf{s}_i$ and $\mathbf{s}_j$, $\rho$ denotes the characteristic length-scale parameter. The hyperparameters $\bm{\uptheta} = (\sigma, \rho)$ are identified through the maximum likelihood estimation (MLE) \cite{Bishop2006Pattern}. In the next section, the performance of the proposed scheme is demonstrated through experiments.
%It is noted that the random vector $\mathbf{Y(s)}$  is represented as 1-dimensional for the ease of notation. However, the generalized format for multi-dimensional outputs is clear.

\section{Experimental Results}
\label{sec: Simulation Results}
% The performance of the proposed coverage sampling planner is evaluated using a real-world dataset by numerically experiments in this section. 
This section presents the simulation and field tests by implementing the proposed method in field estimation and mapping. The proposed approach is also compared with the existing methods for coverage sampling design and path planning, including square grid-based spanning tree coverage (SGSTC) planner \cite{kapoutsis2017darp}, discrete monotone polygonal partitioning-based (DMPP) planner \cite{wilson2017adaptive}, and hexagonal grid-based TSP (HGTSP) planner.
% Given a power budget, the coverage paths are planned with sampling points distributed on it. Data samples are acquired at the observed locations and then utilized to estimate the underlying field model and map the overall scalar field. 
The SGSTC method was originally designed for sweep coverage planning, which is adopted to coverage sampling for comparison in the experiment. The DMPP method, a variant of the Boustrophedon survey was designed to generate coverage path for sampling and field mapping. The HGTSP method utilizes the same sampling design as the present work but they plan the coverage path by a TSP solver.
%All these planners can generate coverage paths at different resolutions to cover the spatial field. To fairly compare the estimation performance on mapping, they are tuned to find the optimal spacing density that they can achieve for a specific power supply. 
Note that in these methods different parameters are used to adjust the coverage density of sampling. Specifically, the edge length of a fine square cell is set to control the spacing in SGSTC. The distance between the spaced transects determines the spacing in DMPP. The edge length of a hexagonal cell defines the coverage density in HGTSP and HGC. These parameters are defined as the sampling density regulator (SDR) in the present experiments. Consequently, the optimal SDR (OSDR) is the corresponding parameter $l^*$ in Equation (5).

\subsection{Simulation}
\subsubsection{Real-world dataset}
In the simulation, a ground truth map over a surveillance area is chosen from a real-world dataset, NOAA operational model archive and distribution system (NOMADS)  \cite{NOAA2018Noaa} provided by the National Oceanic and Atmospheric Administration (NOAA). The dataset records area measurements of surface salinity of the Caribbean Sea using a radiometer. In the dataset, an estuary area (latitude: 28.4339N-30.4155N, longitude: 87.6968W-89.7512W) with apparent spatial variation is chosen as the study area to evaluate the performance of the planners. A ground truth map is shown in Fig. 5(a), where the colors indicate the surface salinity concentrations in units of psu. % The GPS coordinates are translated into metric coordinates in units of km, as ${\bf{s}} = (s_1, s_2)$, $s_1\in [0,51]$, $s_2\in[0,56]$.

\subsubsection{Mapping performance}
The mapping performance is evaluated by comparing the prediction values with the ground truth values at the unobserved locations. The root mean square error (RMSE) and the average Kriging variance (AKV) are utilized as measures to determine the mapping performance, which are defined as:
\begin{equation}
e_{RMS} = \sqrt{\dfrac{\Sigma_{\mathbf{i}\in I}[\hat{Y}(\mathbf{i})|\mathbf{y}-\hat y(\mathbf{i})]^2}{|I|}},
\bar{\Sigma}_{K} = \dfrac{\Sigma_{\mathbf{i}\in I}\mathbf{\Sigma}^2(\mathbf{i})|\mathbf{y}}{|I|},
\end{equation}
where $I$ denotes the set of grid nodes for spatial interpolation, $\mathbf{i}\in I\subset A$, $|I|$ denotes the element number of the set, and $y(\mathbf{i})$ denotes the ground truth value at locations $\mathbf{i}\in I$. A smaller RMSE indicates a more accurate prediction at the unobserved locations of the field. A smaller AKV indicates a lower estimation uncertainty.

The power supply budget determines the total distance a mobile robot can travel. For simplicity, the budget is designated in a unit of length rather than energy in the simulation. To investigate the mapping performance with respect to the power constraint, different energy budgets are assigned when planning the sampling paths. Fig. 4 shows the simulation results of the compared methods with respect to the different energy budgets. For brevity, some OSDR and field mapping results are presented in Table I and displayed in Fig. 5. The configuration of the binary search for the OSDR is set as $d\in [2, \min (s_1,s_2)/4]$, $s_1\in [0,51]$, $s_2\in[0,56]$, and $\delta = 20$, $e_M = 1$.

	\begin{figure}[h!]
      \centering
      \subfigure[RMSE.]{\label{fig:simulation_rmse}  
      \includegraphics[width=0.23\textwidth]{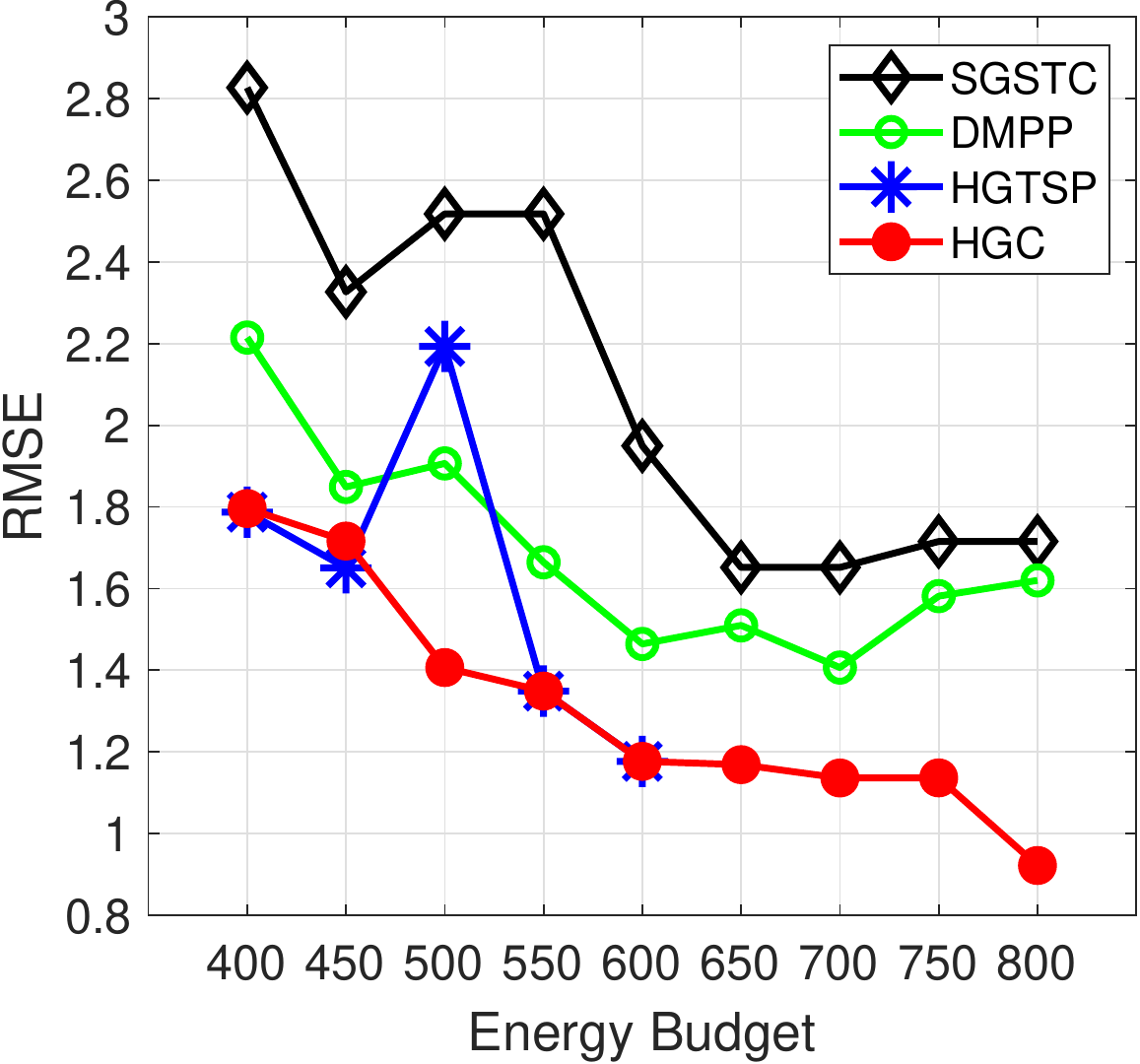}} % \ \ \ \ \ \ \ \
      \subfigure[AKV.]{\label{fig:simulation_akv}  
      \includegraphics[width=0.23\textwidth]{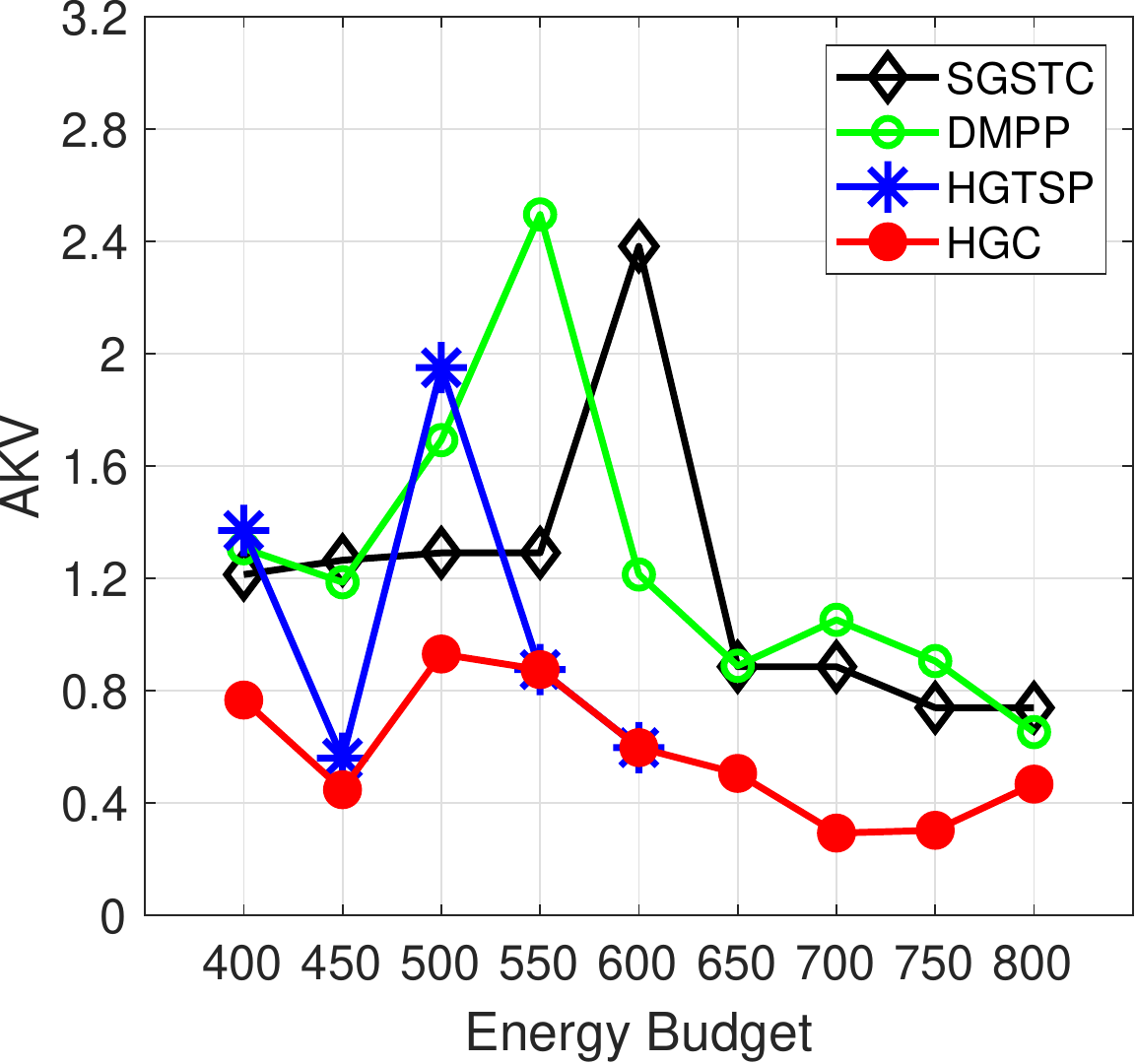}}
      %\subfigure[Ground Truth.]{\label{fig:4a}  
      %\includegraphics[height=4.3cm]{MKVBinary.eps}} 
      %\subfigure[SGSTC.]{\label{fig:4b}  
      %\includegraphics[height=4.3cm]{MKVBruteForce.eps}}

      \caption{Mapping performance on NOMADS dataset.}
      \label{fig:simulation_prediction}
   \end{figure}
	\begin{table}[h!]
	\renewcommand{\arraystretch}{1.05}
    \caption{Mapping performance on NOMADS dataset.} \label{tab:1}
    \centering
\begin{tabular}{ | p{0.8cm} | b{0.9cm} || p{0.9cm} | p{0.9cm} | p{0.9cm} | p{0.8cm} |}
      \hline
      \centering
      \textbf{Budget} & \centering \textbf{Metric} & \textbf{SGSTC } & \textbf{DMPP } & \textbf{HGTSP} & \textbf{HGC}
      \\
      \hline\hline
      \centering & \centering $l^*$ & 5.36 & 5.52 & 4.74 & 4.58 \\
      \centering 400 & \vspace{-2pt} \centering $e_{RMS}$ \vspace{0.1bp} & 2.8272 & 2.2149 & \cellcolor[gray]{.8} 1.7870 & 1.7960 \\
      & \centering $\bar{\Sigma}_{K}$ & 1.2134 & 1.3072 & 1.3704 & \cellcolor[gray]{.8} 0.7662
      \\ \hline
      \centering & \centering $l^*$ & 4.51 & 4.84 & 3.78 & 3.68  \\
      \centering 500 & \vspace{-2pt} \centering $e_{RMS}$ \vspace{0.1bp} & 2.5179 & 1.9065 & 2.1930 & \cellcolor[gray]{.8} 1.4064 \\
      & \centering $\bar{\Sigma}_{K}$ & 1.2907 & 1.6925 & 1.9507 & \cellcolor[gray]{.8} 0.9302
      \\ \hline
      \centering & \centering $l^*$ & 4.05 & 4.16 & 3.10 & 3.10 \\
      \centering 600 & \vspace{-2pt} \centering $e_{RMS}$ \vspace{0.1bp} & 1.9496 & 1.4642 & \cellcolor[gray]{.8} 1.1765 & \cellcolor[gray]{.8} 1.1765 \\
      & \centering $\bar{\Sigma}_{K}$ & 2.3832 & 1.2136 & \cellcolor[gray]{.8} 0.5967 & \cellcolor[gray]{.8} 0.5967
      \\ \hline
      \centering & \centering $l^*$ & 3.76 & 3.68 & N/A & 2.91 \\
      \centering 700 & \vspace{-2pt} \centering $e_{RMS}$ \vspace{0.1bp} & 1.6517 & 1.4064 & N/A & \cellcolor[gray]{.8} 1.1358 \\
      & \centering $\bar{\Sigma}_{K}$ & 0.8850 & 1.0521 & N/A & \cellcolor[gray]{.8} 0.2921
      \\ \hline
      \centering & \centering $l^*$ & 3.25 & 3.35 & N/A & 2.51 \\
      \centering 800 & \vspace{-2pt} \centering $e_{RMS}$ \vspace{0.1bp} & 1.7153 & 1.6199 & N/A & \cellcolor[gray]{.8} 0.9214 \\
      & \centering $\bar{\Sigma}_{K}$ & 0.7389 & 0.6521 & N/A & \cellcolor[gray]{.8} 0.4666
      \\ \hline
    \end{tabular} 
    \begin{flushleft}
    \ \ \ \ \ N/A: Execution did not complete within the time limit of 15 min.
	\\
	\ \ \ \ \ Grey color highlights the best prediction results.    
    \end{flushleft}

  \end{table}

	\begin{figure*}[!t]
      \centering
      \subfigure[Ground Truth.]{\label{fig:6a}  
      \includegraphics[width=0.18\textwidth]{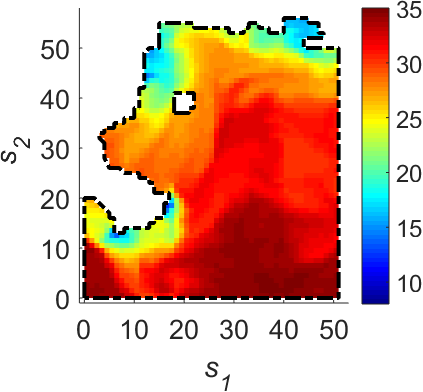}}
      \subfigure[SGSTC (RMSE).]{\label{fig:6b}  
      \includegraphics[width=0.18\textwidth]{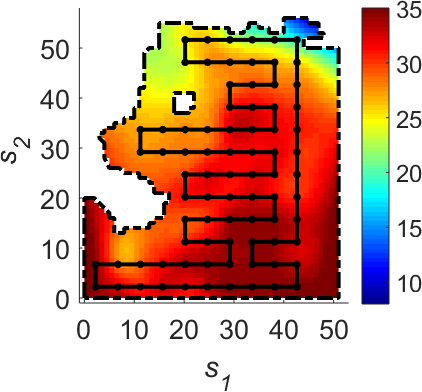}}
      \subfigure[DMPP (RMSE).]{\label{fig:6c}  
      \includegraphics[width=0.18\textwidth]{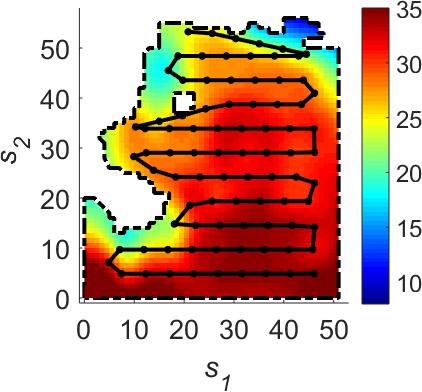}}
      %\subfigure[HGTSP.]{\label{fig:4d}  
      %\includegraphics[width=0.18\textwidth]{HGTSPKrigingResult.eps}}
      \subfigure[HGTSP (RMSE).]{\label{fig:6d}  
      \includegraphics[width=0.18\textwidth]{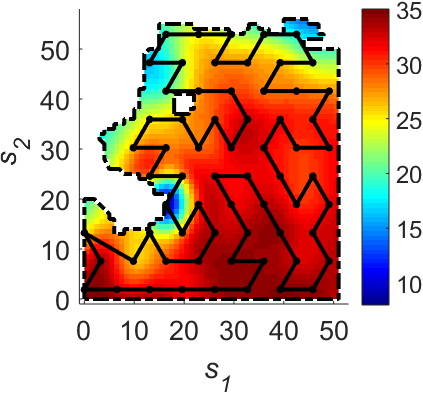}}
      \subfigure[HGC (RMSE).]{\label{fig:6e}
      \includegraphics[width=0.18\textwidth]{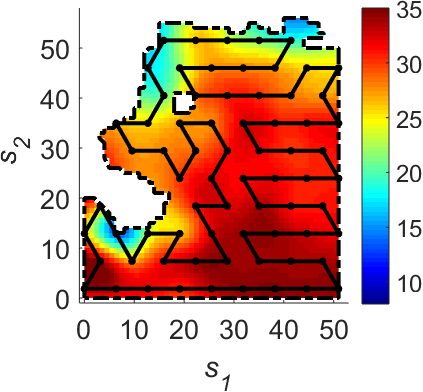}}

      \subfigure[SGSTC (AKV).]{\label{fig:6f}  
      \includegraphics[width=0.18\textwidth]{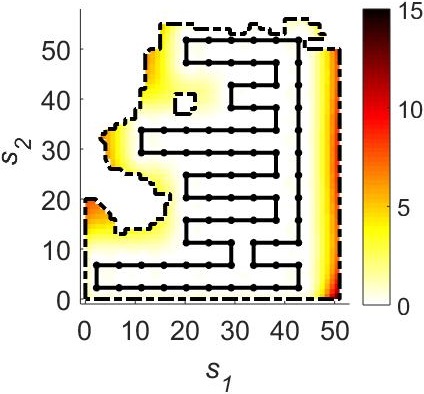}}
      \subfigure[DMPP (AKV).]{\label{fig:6g}  
      \includegraphics[width=0.18\textwidth]{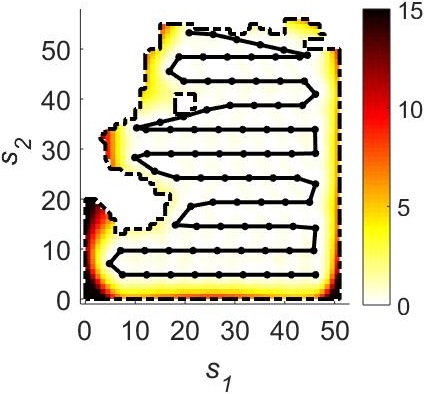}}
      \subfigure[HGTSP (AKV).]{\label{fig:6h}  
      \includegraphics[width=0.18\textwidth]{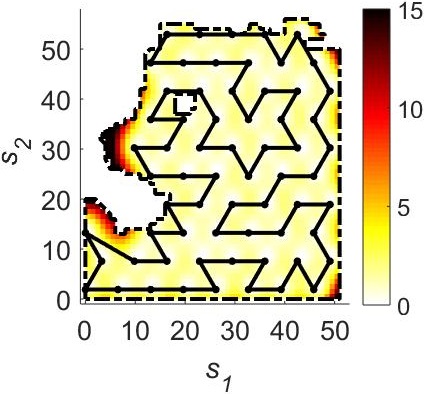}}
      \subfigure[HGC (AKV).]{\label{fig:6i}  
      \includegraphics[width=0.18\textwidth]{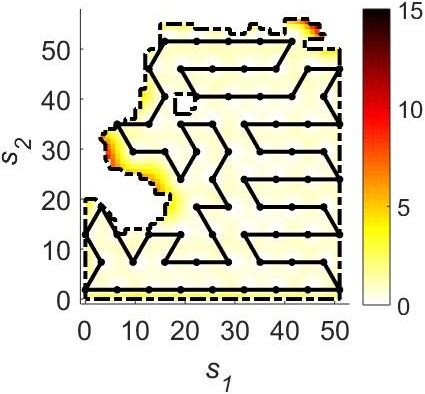}}

\caption{Simulation results of sampling locations and field mapping using different sampling planners (with a budget of 500).} \label{fig:6}
% Energy budget was set at 500. The solid black dots and lines denote the planned SLoIs and sampling paths, respectively. The colors indicate the corresponding predicted value at unobserved locations in the study area.
   \end{figure*}

As clear from Fig. 4 and Table I, the proposed method provides more accurate results of field estimation and mapping when compared with the other approaches, as expressed by the RMSE and the AKV results with different power supply budgets. In the limited number of simulation results, the HGTSP method may provide the lowest RMSE.
% since it is because the TSP solver provides an optimal solution for the planning problem but it is NP hard to solve. 
However, for a larger sampling density, the HGTSP methods were unable to find a plan due to its NP-hard solution. %The HGSTC method in our past work shared the same sampling design and consequently generated similar estimation results as the proposed method. The difference between the HGSTC and the proposed method is mainly on the computational complexity and is demonstrated in the Section \ref{sec: computational cost}.
In Fig. 5, the figures display the sampled locations, the generated coverage path, and the mapping results by making use of the observations at the sampled sites. The solid black dots and lines denote the planned SLoIs and the sampling paths, respectively. The colors in Fig. 5(b)-(e) indicate the corresponding predicted values at the unobserved locations in the study area. The colors in Fig. 5(f)-(i) display the corresponding estimated Kriging variances at the unobserved locations, which indicate estimation uncertainty in the monitored field. Fig. 5 demonstrate that the HGC method outperforms the other sampling planners on field estimation and mapping. % by generating the lowest Kriging variance quantities in the study area.

% In real applications, if a robot can not follow the planned path properly, the worst-case energy cost can be used.

\subsection{Field Tests}
The proposed planner was further evaluated by physical field tests. It was implemented on a UAV-enabled mobile sensor for aquatic environmental monitoring. The mobile sensor was built upon a DJI Phantom 4 quadrotor with a Libelium Waspmote board and a Raspberry Pi 3 microcomputer mounted on the UAV for data acquisition and onboard processing, respectively. A conductivity probe was equipped to measure electrical conductivity of water source in an in-situ manner, which connected to the data acquisition board to provide scalar readings with units of $\mu$S/cm. The system has been deployed at the Yosef Wosk Reflecting Pool, which is a natural pool at University of British Columbia, Canada. The UAV-enabled mobile sensor and the study area of the monitored aquatic environment are shown in Fig. 6.

      \begin{figure}[!t]
      \centering
      %\subfigure[]{\label{fig:10}  
      %\includegraphics[width=0.4\textwidth]{Figure10/IMG_2815.JPG}}
            %\subfigure[Aerial photograph of the monitored environment.]{\label{fig:10b}  
      % \includegraphics[width=0.4\textwidth]{Figure10/Pool.png}%}
      
      \includegraphics[width=0.45\textwidth]{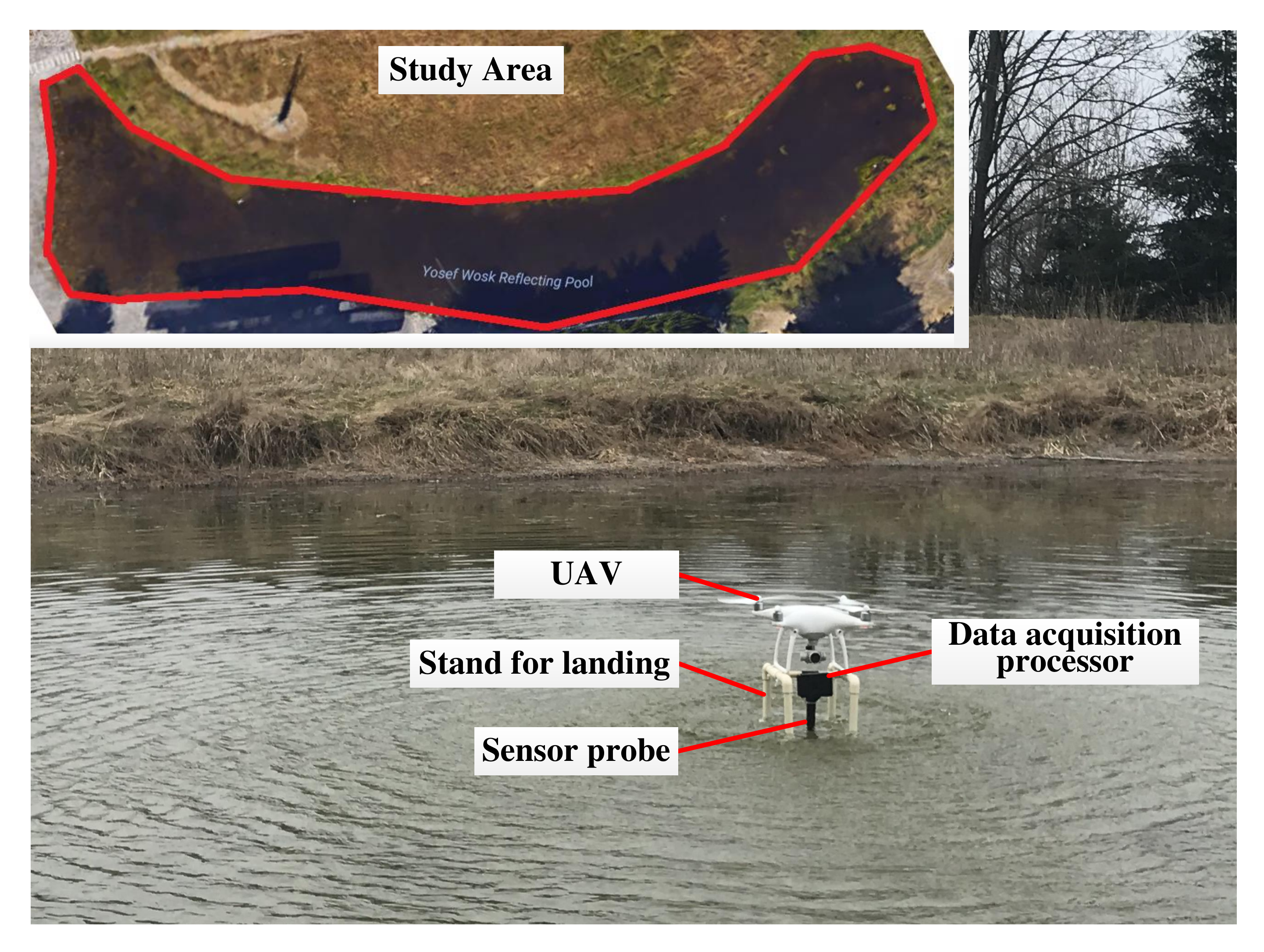}
      \caption{Study area and field deployment of the UAV-enabled mobile sensor for water quality monitoring.}
      \label{fig: field deployment}
   \end{figure}

% In mobile sensing practice, propelling is the primary factor of energy consumption. Other cost sources such as sensing, on-board data processing, and wireless communication are trivial compared with the energy cost of propelling for mobility. % For the developed system, a fully charged 14.8 V, 5200 mA$\cdot$h LiPo battery can enable the mobile robot to run continuously around 30-40 min (depending on the hydrological dynamics of the monitored environment) at the average speed of 0.4 m/s. The additional system settings for the field tests are given in Table III.
   % \input{Table/Table3}
   
In each field test, one UAV was implemented by applying one of the planners to carry out the corresponding sampling mission. Meanwhile, another UAV was deployed to collect data at some unobserved locations, which was used as a validation dataset to evaluate the performance of field estimation and mapping. The fully charged battery (81.3 Wh) enabled the USV to fly for roughly 20 minutes at the average speed of 1 m/s. The hovering time for carrying out a measurement at each target location was set to 10 seconds (energy cost of about 2 Wh). The mapping results of the different planners were based on the underlying environmental models that were learned by the observations from the first UAV. The measurements taken by the second UAV were utilized as the ground truth to validate the mapping results.

% For each experiment, fully charged batteries were plugged into both two robots. One-third of the battery capacity was assigned to complete the sampling missions, which means the total travel distance was approximately around 320 m. One-half of the battery capacity was assigned to the second mobile robot to collect validation data samples. The sensor mote in the robot for operating the sampling mission was updated by around 0.1 Hz, which formed the sampling interval distance to be around 3.5-4 m. In comparison, the sensors were measured by around 0.2 Hz to collect denser data samples over the surface water, which made the sampling resolution at about 2 m.
%Electrical conductivity were selected as the monitored parameters to evaluate the estimation performance. For each field trial, the underlying environmental models of the monitored parameters were learned by the observations from the first robot. The measurements taken by the second robot were utilized to validate the performance of field estimation and mapping. 

      \begin{figure*}[!t]
      \centering
      % \subfigure[]{\label{fig:6}
      % \includegraphics[width=0.4\textwidth ]{Figure11/Fig11b.eps}}
      % \subfigure[]{\label{fig:10a}
      % \includegraphics[width=0.4\textwidth]{Figure11/Fig11c.eps}} \\
      % \subfigure[]{\label{fig:6}
      % \includegraphics[width=0.4\textwidth]{Figure11/Fig11d.eps}}
            \subfigure[Generated coverage path for sampling.]{\label{fig:7a}
      \includegraphics[width=0.30\textwidth]{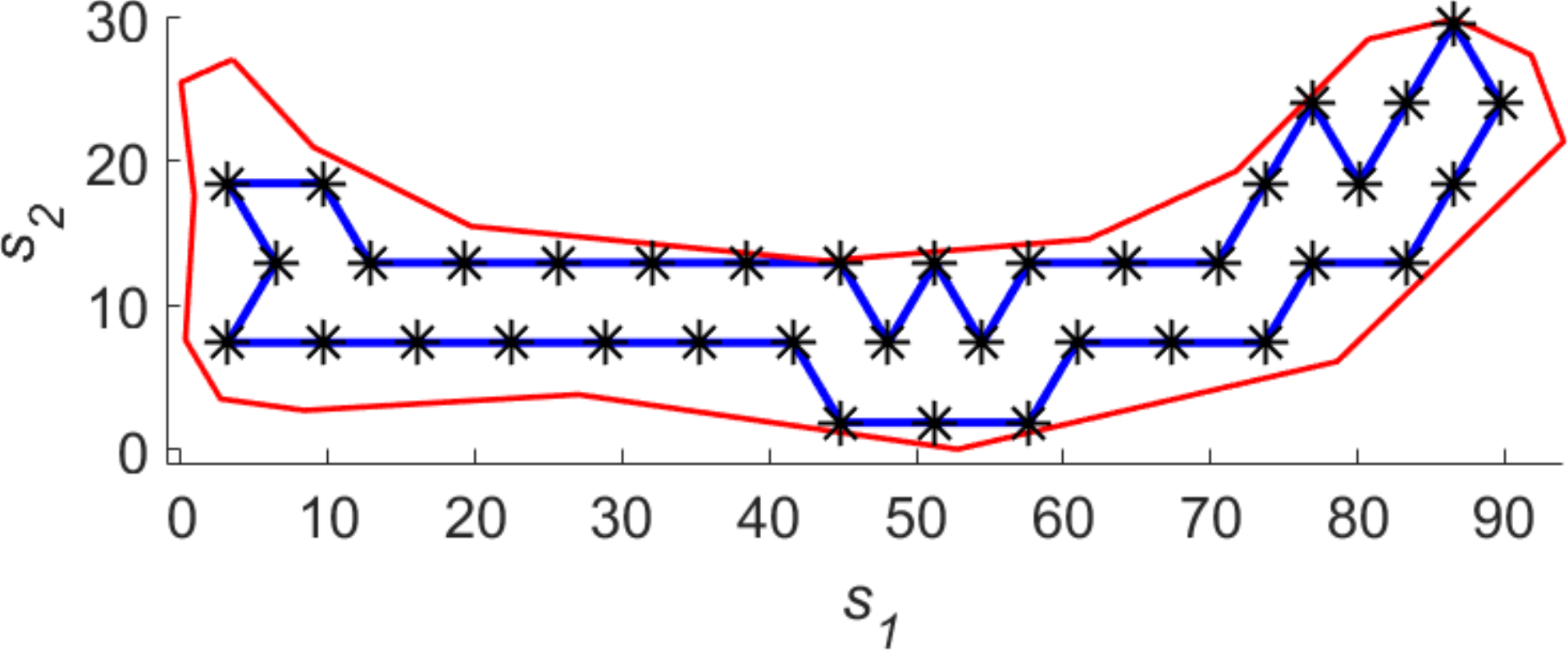}}
      		\subfigure[Sampled locations and conductivity map.]{\label{fig:7b}
      \includegraphics[width=0.32\textwidth]{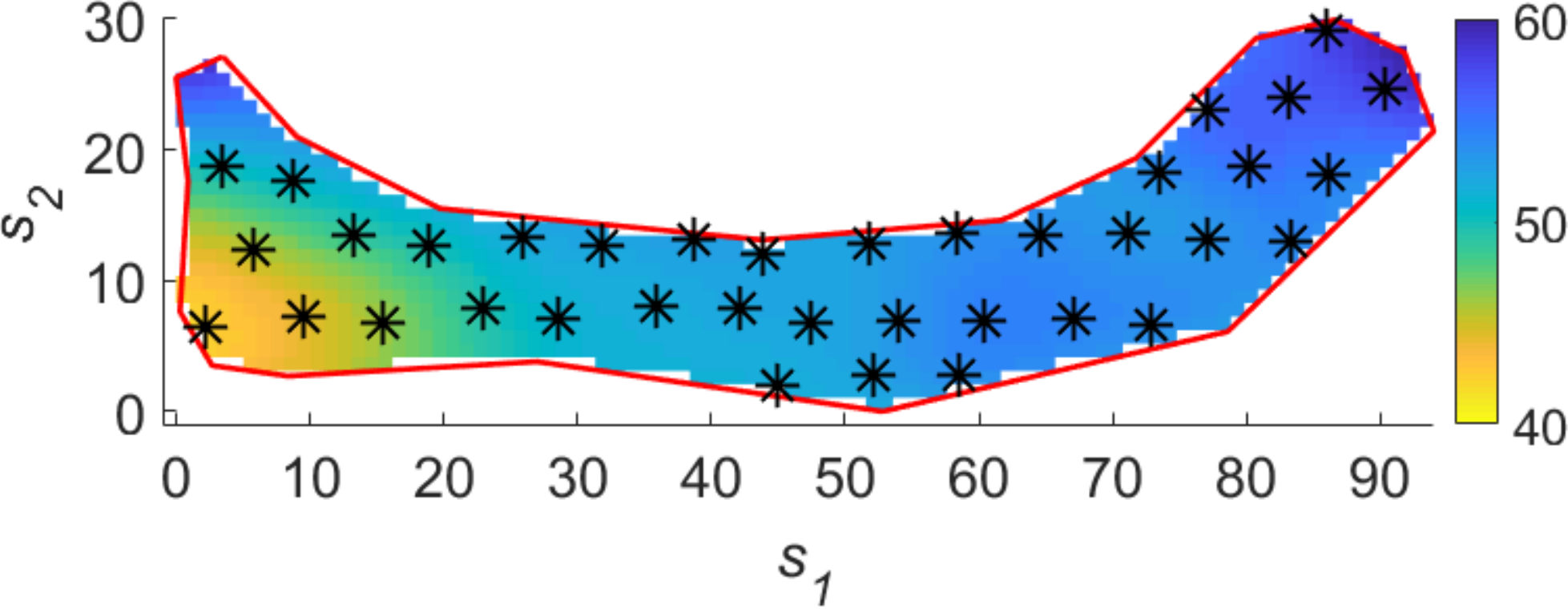}}
      		\subfigure[Sampled locations and uncertainty map.]{\label{fig:7c}
      \includegraphics[width=0.32\textwidth]{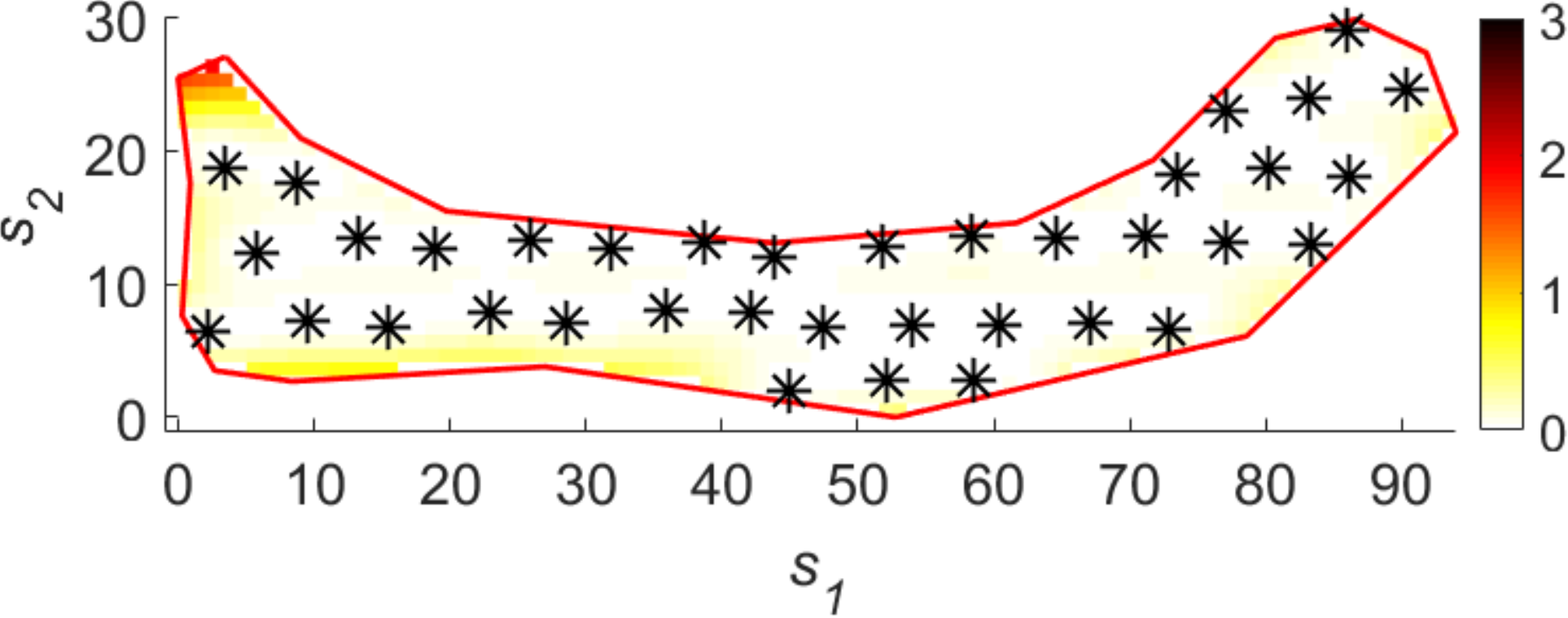}}
      \vspace{-5pt}
      \caption{%(a) Generated coverage path for sampling of the proposed method. The solid red and blue contour denote the boundary of the study area and the planned path, respectively. (b) Sampling locations in the field. The blue star markers denote the sampled locations. The black dot markers denote the sampled locations for validation. (c) Temperature mapping result. (d) 
      Field test results using the proposed HGC sampling planner.}
      \label{figurelabel}
   \end{figure*}  

Table II shows the performance of the RMSE  and AKV by implementing the proposed and compared planners. % In the table, all the RMSE results are relatively small. It is because the monitored pool has a limited spatial scale and display unapparent spatial variation during experiments. 
The proposed planner demonstrated its superior prediction performance compared to the other planners. To further display the mapping results, Fig. 7 provides the planning, prediction, and estimation results at the monitored pool by making use of the observations from the generated HGC sampling planner. The experimental results of the field tests further validated the conclusion in the numerical simulation that the proposed planner outperforms the state-of-the-art methods on environmental field estimation and mapping.

   \begin{comment}
\begin{figure*}[!t]
  %\vspace*{\floatsep} 
  \begin{minipage}{.4\linewidth}
%\begin{table}[!t]
    \centering
    \captionof{table}{Algorithm Performance of Field Tests}\label{tbl: Algorithm Performance on Sampling Path Planning Approaches}
    \begin{tabular}{ | p{1cm} || p{1cm} | p{1cm} | p{1cm} | p{0.9cm} |}
      \hline
      \centering
      \textbf{Metric} & \textbf{SGSTC} & \textbf{DMPP} & \textbf{RSTSP} & \textbf{HGC} \\
      \hline \hline
      \centering \text{RMSE} & \small 0.1545 & \small 0.1429 & \small 0.1282 & \cellcolor[gray]{.8} \small 0.1277 \\ %\cellcolor[gray]{.8} 0.1277
      \hline
      \centering \text{AKV} & \small 0.1545 & \small 0.1429 & \small 0.1282 & \cellcolor[gray]{.8} \small 0.1277 \\
      \hline
      %%%%%%%%%%%%% end 6th row
    \end{tabular}
  %\end{table}
  \end{minipage}%
  \begin{minipage}{0.6\linewidth}
        \subfigure[RMSE.]{\label{fig:6d}  
    \centering \includegraphics[width=0.5\textwidth]{Figure11/Fig11e.eps}}
            \subfigure[AKV.]{\label{fig:6d}  
    \centering \includegraphics[width=0.5\textwidth]{Figure11/Fig11e.eps}}
    \captionof{figure}{Filed mapping results of the proposed HGC sampling planner.}
  \end{minipage}
\end{figure*}%
 \end{comment}

\begin{table}[!t]
  \renewcommand{\arraystretch}{1.2}
  \caption{Planner performance of field tests.}\label{tbl: Algorithm Performance on Sampling Path Planning Approaches}
    \centering
    \begin{tabular}{ | p{1cm} || p{0.9cm} | p{0.8cm} | p{0.9cm} | p{0.8cm} |}
      \hline
      \centering
      \textbf{Metric} & \textbf{SGSTC} & \textbf{DMPP} & \textbf{HGTSP} & \textbf{HGC} \\
      \hline \hline
      \centering $e_{RMS}$ \vspace{0.1bp} & 2.5155 & 1.0760 & 0.6821 & \cellcolor[gray]{.8} 0.6754 \\ %\cellcolor[gray]{.8} 0.1277
      \hline
	  \centering $\bar\Sigma_K$ & 1.4196 & 0.8083 & 0.1961 & \cellcolor[gray]{.8} 0.1072 \\
      \hline
      %%%%%%%%%%%%% end 6th row
    \end{tabular}
  \end{table}

% Fig. 8 shows more information of the generated coverage path for sampling, the observed sampling locations and validation sites, and the mapping results by making use of the observations from the proposed sampling operation in the field operation.

\section{CONCLUSIONS}
In the present paper, a hexagonal grid-based coverage sampling planner was proposed for spatial exploration, random field estimation, and environmental mapping. The proposed planning strategies provided a reliable and efficient coverage sampling mission for USV-enabled mobile sensing with a in-situ sensor, to make an effective prior survey of an unknown environment under a power constraint. The proposed planner could distribute coverage sampling densely and evenly across the overall study area while generating the corresponding coverage path to visit the target sampling sites. The experimental results on the real-world dataset and field tests validated the planner performance on mapping accuracy. In practical applications, the proposed scheme satisfies many circumstances in environmental monitoring, such as building an field map of an unknown environment, scheduling an initial deployment to gather sufficient useful information that can be exploited for further implementation or long-term monitoring. In future work, since the planned sampling path forms a path cycle, it can be divided into multiple optimal sub-paths to schedule multiple robotic sensors for sensing according to their power supply conditions.



% \section*{ACKNOWLEDGMENT}
% This work has supported by the research grant held by Prof. Clarence W. de Silva through the India-Canada Centre for Innovative Multidisciplinary Partnership to accelerate Community Transformation and Sustainability (Grant No. 11R18083).

\bibliographystyle{IEEEtran}
% \bibliography{IEEEabrv,../bib/paper}
\bibliography{mybibfile}
\end{document}